\documentclass{article}

\usepackage{microtype}
\usepackage{graphicx}
\usepackage{float} 
\usepackage{placeins} 
\usepackage{subfigure}
\usepackage{booktabs}

\usepackage{hyperref}
\usepackage{bbm} 
\usepackage{tikz}
\usepackage{pgfplots}
\usepgfplotslibrary{groupplots}
\pgfplotsset{compat=1.17}
\usepgfplotslibrary{fillbetween}
\usepackage{xcolor}

\usepackage[accepted]{icml2026}

\usepackage{amsmath}
\usepackage{amssymb}
\usepackage{mathtools}
\usepackage{amsthm}
\usepackage[capitalize,noabbrev]{cleveref}
\usepackage{mdframed}

\theoremstyle{plain}
\newtheorem{theorem}{Theorem}[section]
\newtheorem{proposition}[theorem]{Proposition}

\theoremstyle{definition}

\theoremstyle{remark}

\newmdenv[
  linewidth=0.8pt,
  roundcorner=4pt,
  linecolor=black!30,
  backgroundcolor=black!3,
  innertopmargin=6pt,
  innerbottommargin=6pt,
  innerleftmargin=8pt,
  innerrightmargin=8pt
]{thmframe}

\usepackage[textsize=tiny]{todonotes}

\icmltitlerunning{FLIP for VLMs}

\begin{document}

\twocolumn[
\icmltitle{FLIP: Final Layer Inference-Time Probing for Vision--Language Models}


\begin{icmlauthorlist}
\icmlauthor{Drandreb Earl O. Juanico}{xxx}
\icmlauthor{Rowel O. Atienza}{xxx,yyy}
\end{icmlauthorlist}

\icmlaffiliation{xxx}{AI Graduate Program, University of the Philippines, Diliman, Quezon City, Philippines 1101}
\icmlaffiliation{yyy}{EEE Institute, College of Engineering, University of the Philippines, Diliman, Quezon City, Philippines 1101}

\icmlcorrespondingauthor{Drandreb Earl O. Juanico}{dojuanico@alum.up.edu.ph}

\icmlkeywords{Methods, Benchmarking interpretability, Feature geometry}

\vskip 0.3in
]

\printAffiliationsAndNotice{\icmlEqualContribution}


\newcommand{\Iflip}{\mathbbm{1}_{\mathrm{FLIP}}}
\newcommand{\TPfifty}{\mathrm{TP}_{50}}
\newcommand{\Rfifty}{R_{50}}
\newcommand{\dRfifty}{\Delta R_{50}}
\newcommand{\dAcc}{\Delta\mathrm{Acc}}
\newcommand{\ecount}{\mathcal{E}_{\mathrm{count}}}

\begin{abstract}
We present FLIP, a final-layer inference-time probe for testing whether a
logit-facing intervention site in an open-weight vision--language model (VLM) supports
structured, task-linked computation rather than generic perturbation. Behavioral
change under internal intervention is otherwise mechanistically ambiguous: it
may reflect improved use of visual evidence, generic output instability, or
outright degradation. FLIP applies elementwise flooring to the final normalized
hidden state before logit computation, leaving parameters, prompts, and decoding
unchanged. On a controlled detection/counting probe, sweeping intervention
strength reveals three regions: negligible change, a bounded interior regime in
which detection recall at IoU $0.50$ ($R_{50}$) improves while tolerant counting
error ($\mathcal{E}_{\mathrm{count}}$) falls, and over-suppression. We formalize a four-criterion
probe-and-sweep protocol for disciplining the interpretation of intervention
effects: regime structure, grounding-proxy alignment, feature-coherence
dependence, and failure to reproduce the same positive regime on a
performance-based negative control. The post-normalization state passed to the output head is the logit-facing
instantiation of this test; under a non-targeted flooring sweep it satisfies the
full protocol. Raw decoder-layer interventions---including the last-block output before final
normalization---and the singleton-pair left/right control fail to reproduce the
Final-site signature, while same-site operators and multiple VLMs replicate it. FLIP is therefore a validation step for intervention-based
mechanistic interpretability, not a steering method.
\end{abstract}

\section{Introduction}
\label{sec:introduction}

Open-weight vision-language models expose final-layer hidden states at inference
time, but behavioral change under intervention is mechanistically ambiguous: it
may reflect improved use of visual evidence, generic instability, or outright
degradation. Single-point interventions cannot resolve this ambiguity. We
therefore evaluate intervention sites under sweeps over intervention strength
$\vartheta$ and ask whether they pass a four-criterion probe-and-sweep protocol
for structured, task-linked computation rather than generic perturbation; a
fuller definition of the latter is given in
Appendix~\ref{app:generic_perturbation}.

FLIP (Final Layer Inference-Time Probing) applies elementwise flooring to the
post-normalization Final-site hidden state immediately before logit computation
(\cref{sec:flip}). We use the standard linearity of the output head as a
baseline for interpreting this site: a hidden-state displacement can be useful
only insofar as it projects onto task-relevant output directions. The empirical
question is therefore not whether FLIP can perturb outputs, but whether a
non-targeted flooring sweep reveals a bounded, task-linked response. On a
controlled detection/counting probe, we observe three response regions:
negligible change, an interior regime with improved $\Rfifty$ and reduced
$\ecount$, and over-suppression.

The aim is not to identify a circuit, but to validate whether an intervention
effect is structured enough to justify circuit-level follow-up. FLIP therefore
treats behavioral change as credible only if it survives independent constraints:
regime structure, grounding-proxy alignment, feature-coherence dependence, and
task-specific negative control. Raw decoder-layer checks
(\cref{fig:cross_layer_check}; Appendix~\ref{app:cross_layer_failure},
\Cref{fig:cross_layer_failure_full}) provide site-specific contrasts under the
implemented sweep, while cross-architecture and same-site operator replications
(\cref{fig:multimodel_composite_two_panel_centered};
Appendices~\ref{app:cross_architecture},~\ref{app:operator_baselines}) test the
protocol-level signature beyond one model or operator. The delta-method product
summary (\Cref{tab:mediation-4b}) tests compatibility with a shared
detection-to-counting grounding factor.

\paragraph{Contributions.}
(1) We propose a probe-and-sweep protocol that treats intervention effects as
objects requiring validation, not as self-interpreting performance changes. (2)
We instantiate it with FLIP, a minimal Final-site flooring probe grounded in the
standard output-head baseline. (3) We show that primary-task gains are
insufficient: accepted effects must also satisfy grounding-proxy alignment,
feature-coherence dependence, negative-control contrast, and logit-facing
site specificity. (4) We replicate this protocol-level signature across same-site
magnitude-bounding operators and multiple vision-language models.

\section{Background and Related Work}
\label{sec:related}

Intervention-based interpretability trades off causal specificity and population
coverage. Causal abstraction uses interchange interventions to test aligned
causal variables \cite{geiger2021causal}; causal mediation asks which internal
components are implicated in behavior \cite{vig2020investigating}; activation
patching and automated circuit discovery localize task-relevant components after
a metric and dataset have been chosen
\cite{zhang2023towards,conmy2023towards}; and circuit-tracing methods aim to
recover prompt-level computational graphs \cite{ameisen2025circuit}. These
methods provide stronger mechanistic follow-up, but their conclusions can depend
on the chosen behavior, metric, examples, and ablation design
\cite{miller2024transformer}. Recent VLM work similarly analyzes visual-token
processing, learns interpretable feature bases, or intervenes through latent
and hidden-state steering to improve grounding or reduce hallucination
\cite{neo2024towards,pach2025sparse,yang2026circuit,liu2025reducing,
su2025activation, juanico2025interpretable, li2025hidden}. FLIP targets the preceding selection problem:
before assigning an effect to a circuit or using it for steering, it asks
whether a candidate site exhibits a population-level, task-linked,
control-sensitive signature that warrants mechanistic follow-up.

\begin{figure*}[t]
  \centering
  \begin{tikzpicture}[font=\normalsize]
  \definecolor{barbase}{RGB}{27,79,114}
  \definecolor{barthresh}{RGB}{178,34,34}
  \definecolor{framegray}{RGB}{200,200,200}
  \pgfplotsset{
    baseaxis/.style={
      width=6.2cm, height=3.6cm,
      ymin=-11.5, ymax=8.0, xmin=0.5, xmax=11.5,
      xtick=\empty, ytick={-10,-5,0,5},
      ticklabel style={font=\small, text=black},
      tick style={draw=none}, axis line style={draw=none},
      axis background/.style={fill=white},
      clip=false, enlargelimits=false,
    }
  }
  \def\leftx{0cm}
  \def\rightx{6.5cm}
  \begin{axis}[baseaxis, at={(\leftx,0)}, anchor=south west, yticklabel pos=left, set layers]
    \draw[draw=framegray, line width=0.6pt] (rel axis cs:0,0) rectangle (rel axis cs:1,1);
    \addplot+[ybar, bar width=8pt, draw=barbase, fill=barbase, mark=none] coordinates {
      (1,0.2)(2,1.2)(3,-11)(4,-3.4)(5,0.1)(6,-9)(7,-0.75)(8,-1)(9,3)(10,7)(11,2.5)
    };
    \addplot+[ybar, bar width=8pt, draw=barthresh, fill opacity=0, ultra thick, mark=none] coordinates {(3,-11)};
    \addplot+[ybar, bar width=8pt, draw=barthresh, fill opacity=0, ultra thick, mark=none] coordinates {(4,-3.4)};
    \addplot+[ybar, bar width=8pt, draw=barthresh, fill opacity=0, ultra thick, mark=none] coordinates {(6,-9)};
    \addplot+[barthresh, ultra thick, mark=none] coordinates {(0.5,-2)(11.5,-2)};
    \node[anchor=east, font=\normalsize, text=barthresh, xshift=-18pt] (varthetaLabel) at (axis cs:0.3,-2) {$\vartheta=-2$};
  \end{axis}
  \begin{axis}[baseaxis, at={(\rightx,0)}, anchor=south west, yticklabels={}, set layers]
    \draw[draw=framegray, line width=0.6pt] (rel axis cs:0,0) rectangle (rel axis cs:1,1);
    \addplot+[ybar, bar width=8pt, draw=barbase, fill=barbase, mark=none] coordinates {
      (1,0.2)(2,1.2)(3,-2)(4,-2)(5,0.1)(6,-2)(7,-0.75)(8,-1)(9,3)(10,7)(11,2.5)
    };
    \addplot+[ybar, bar width=8pt, draw=barthresh, fill opacity=0, ultra thick, mark=none] coordinates {(3,-2)};
    \addplot+[ybar, bar width=8pt, draw=barthresh, fill opacity=0, ultra thick, mark=none] coordinates {(4,-2)};
    \addplot+[ybar, bar width=8pt, draw=barthresh, fill opacity=0, ultra thick, mark=none] coordinates {(6,-2)};
    \addplot+[barthresh, ultra thick, mark=none] coordinates {(0.5,-2)(11.5,-2)};
  \end{axis}
  \draw[->, line width=1.2pt] (4.7cm,1.45cm) -- (6.4cm,1.45cm);
  \node[font=\normalsize] at (5.5cm,1.9cm) {FLIP};
  \node[font=\scriptsize, anchor=north west] at ($(varthetaLabel.west)+(1.0cm,-1.25cm)$)
    {$\mathbf{z}=\langle 0.2,\!1.2,\!-11,\!-3.4,\!0.1,\!-9,\!-0.75,\!-1,\!3,\!7,\!2.5\rangle$};
  \node[font=\scriptsize, anchor=north west] at (5.5cm,-0.15cm)
    {$\tilde{\mathbf{z}}=\langle 0.2,\!1.2,\!{\color{barthresh}\boxed{-2}},\!{\color{barthresh}\boxed{-2}},\!0.1,\!{\color{barthresh}\boxed{-2}},\,-0.75,\!-1,\!3,\!7,\!2.5\rangle$};
  \end{tikzpicture}
\caption{\textbf{Hidden-state flooring view of FLIP.}
Components below $\vartheta$ are clamped elementwise at the post-normalization
Final site before logit computation (here, indices 3, 4, and 6). FLIP uses
$\widetilde{\mathbf z}=\max(\mathbf z,\vartheta)$ as the intervention; attention
and rollout maps are used only as downstream diagnostics.}
  \label{fig:hidden_state_flooring}
\end{figure*}
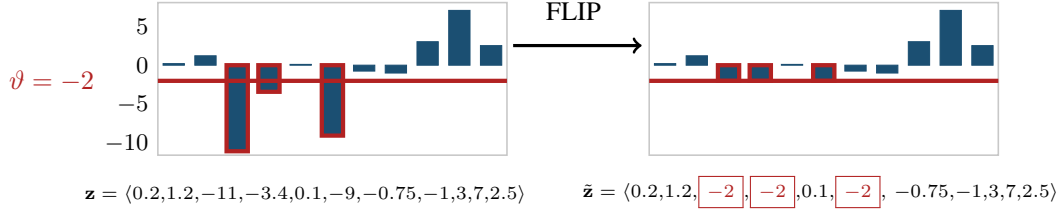

\section{FLIP: Final Layer Inference-Time Probing}
\label{sec:flip}

FLIP is an inference-time intervention applied to the post-normalization final
hidden-state tensor. \Cref{fig:hidden_state_flooring} shows the operator:
components below $\vartheta$ are clamped elementwise before logit computation,
while all others are left unchanged.

\subsection{Operational Definition}
\label{sec:flip-flooring-def}

Let $\mathbf{z}^{(t)}\in\mathbb{R}^{n_t\times d}$ denote the post-normalization
Final-site tensor passed to the output head during generation call $t$, with
$n_t$ equal to the prompt/image prefill length for the first next-token logits and
$n_t=1$ for cached decoding steps. For strength $\vartheta\in\mathbb{R}$, FLIP
outputs
\begin{equation}
    \widetilde{\mathbf{z}}^{(t)}=\max(\mathbf{z}^{(t)},\vartheta),
\end{equation}
element-wise, the replacement is applied at every next-token logit computation
during generation. Thus, FLIP covers the prefill logits used for the first response
token and each subsequent autoregressive decoding step, while leaving weights,
prompts, tokenization, decoding rules, and KV caches unchanged.

\begin{thmframe}
\noindent\textbf{FLIP (definition used in all experiments).}
Given the post-normalization Final-site hidden state $\mathbf{z}$ and strength
$\vartheta$, FLIP outputs
$\widetilde{\mathbf{z}}=\max(\mathbf{z},\vartheta)$ element-wise before logit
computation. No model weights, prompts, or decoding rules are changed.
\end{thmframe}

In the main experiments, FLIP targets the post-normalization final hidden state
passed to the output head. We distinguish this \emph{Final} site from raw
decoder-layer outputs, including the pre-normalization last-block output. FLIP is
a probe, not an optimization method.

\section{Tasks and Vision-Centric Evaluation Setting}
\label{sec:tasks}

Unless otherwise stated, main-text experiments use Qwen3-VL-4B~\cite{bai2025qwen3vl}; additional model
results are reported in the appendix.

\subsection{Intermediate Visual Signals}
\label{sec:ivs}

\paragraph{Prompted localization.}
We query the model with \texttt{\small Detect <object>. Provide correct bounding boxes.}
and parse outputs into a common pixel-space representation; model-specific
details are given in Appendix~\ref{app:matching}.

\paragraph{Dataset and supervision.}
Localization prompts are instantiated on a clustered subset of
MS~COCO val2017~\cite{lin2014microsoft} comprising 7{,}056 image--prompt pairs,
constructed image-centrically with one query per present object category.
Image-level clustering is used because prompts drawn from the same image are not
independent; annotations are used only to instantiate \texttt{<object>} and score
outputs.

\subsection{Reasoning-Based Counting}
\label{sec:count}

\paragraph{Counting prompt.}
\texttt{\small How many <object>? Answer with integer only.} isolates errors
attributable to visual instance discrimination rather than formatting.

\subsection{Vision-Centric Benchmarks}
\label{sec:bench}

We additionally evaluate MMStar, MindCube, and NaturalBench via LMMS-Eval as
intervention probes rather than optimization targets.\footnote{Third-party asset
credits, licenses, and terms of use for Qwen3-VL, Kimi-VL-A3B, MS COCO val2017,
and LMMS-Eval are documented in the supplemental repository.}

\section{Theory}
\label{sec:theory}

This section gives the simple operator and output-head facts used to motivate
the probe. These facts are not themselves the main novelty: they provide the
baseline against which the empirical protocol is interpreted. Since FLIP is not
trained, task-conditioned, or optimized, any useful effect must come from
alignment between the induced displacement and task-relevant directions already
present in the logit-facing representation.

\subsection{Componentwise Flooring and Perturbation Growth}
\label{sec:A_and_M}

For $x\in\mathbb{R}$ and threshold $\vartheta\in\mathbb{R}$, define
\begin{equation}
\phi_{\vartheta}(x)=\max(x,\vartheta),
\qquad
\delta_{\vartheta}(x)=(\vartheta-x)_+.
\end{equation}
Applied elementwise to $\mathbf{z}\in\mathbb{R}^{n\times d}$,
\begin{equation}
\widetilde{\mathbf{z}}=\phi_{\vartheta}(\mathbf{z}),
\qquad
\mathbf{\Delta}_{\vartheta}=\widetilde{\mathbf{z}}-\mathbf{z}.
\end{equation}

Define the active set
\begin{equation}
\mathcal{S}_{\vartheta}=\{(i,k): z_{ik}<\vartheta\}, \qquad
A(\vartheta)=\frac{|\mathcal{S}_{\vartheta}|}{nd},
\end{equation}
and mean clamp magnitude
\begin{equation}
M(\vartheta)=\frac{1}{nd}\sum_{i,k}(\vartheta-z_{ik})_+.
\end{equation}

\begin{thmframe}
\begin{proposition}[Monotone clamping]
For every $(i,k)$,
\begin{equation}
\widetilde z_{ik}=z_{ik}+(\vartheta-z_{ik})_+,
\end{equation}
and $\|\mathbf{\Delta}_{\vartheta}\|_p$ is non-decreasing in $\vartheta$ for all
$p\in[1,\infty]$.
\label{prop:monotone_clamping}
\end{proposition}
\end{thmframe}

As $\vartheta$ increases, FLIP passes from near-inactivity, to a bounded
intermediate region, to widespread clamping. This geometry defines the strength
axis of the probe but does not by itself determine decoded task metrics.

\subsection{Standard Output-Head Alignment Baseline}

Because FLIP is applied immediately before the output head, its first-order
interpretation follows from the standard linear logit map. For a final-token
hidden state $\mathbf z_t$ with output matrix $W$,
\begin{equation}
\widetilde{\ell}_t-\ell_t
=
W^\top \Delta_{\vartheta,t},
\qquad
\Delta_{\vartheta,t}=(\vartheta-\mathbf z_t)_+ .
\end{equation}
For a task-relevant token set $Y^+$ and contrast set $Y^-$, define the
log-sum-exp margin
\begin{equation}
m(\mathbf z_t)
=
\log\sum_{y\in Y^+}\exp(w_y^\top \mathbf z_t)
-
\log\sum_{y\in Y^-}\exp(w_y^\top \mathbf z_t).
\end{equation}
A first-order expansion gives
\begin{equation}
m(\widetilde{\mathbf z}_t)-m(\mathbf z_t)
\approx
\left\langle
\bar w_{Y^+}(\mathbf z_t)-\bar w_{Y^-}(\mathbf z_t),
\Delta_{\vartheta,t}
\right\rangle ,
\end{equation}
where $\bar w_{Y^+}$ and $\bar w_{Y^-}$ are softmax-weighted output-head
directions over the two token sets.

This calculation is not a new representation theory; it defines the baseline for
interpreting FLIP. Since the clamp is object-agnostic, any useful change must
come from projection onto task-relevant output directions. The protocol then tests whether that necessary condition is accompanied by
grounding-proxy alignment, feature-coherence dependence, logit-facing site
specificity, and negative-control contrast.

\section{Grounding-Proxy Association Analysis}
\label{sec:cca}

This section tests whether FLIP effects are compatible with a grounding-linked
pathway, without claiming full causal identification.

Let $\Iflip\in\{0,1\}$ denote intervention. Detection quality is
\begin{equation}
\Rfifty = \frac{\TPfifty}{\mathrm{gt\_count}}, \qquad
\dRfifty = \Rfifty - \mathbb{E}[\Rfifty \mid \Iflip=0].
\end{equation}
For counting, let $y$ be the parsed model count and
$y^\star=\mathrm{gt\_count}$ the ground-truth count; the tolerant error is
\begin{equation}
\ecount = \max(0, |y-y^\star|-1).
\end{equation}

Localization and counting are elicited by separate prompts and scored from
independent outputs, so $\dRfifty$ is not a literal within-pass mediator of
$\ecount$. We instead use it as an observable proxy for an unobserved
intervention-sensitive grounding state:
\[
\Iflip \rightarrow G_{\vartheta},\qquad
G_{\vartheta}\leadsto \{\dRfifty,\ecount\}.
\]
We estimate the product summary from
\[
\dRfifty \sim \Iflip,\qquad
\ecount \sim \Iflip+\dRfifty.
\]
The first model is a Gaussian GLM with image-clustered SEs and supplies the
$a$-term; the second is a Poisson GLM on strict counting responses and supplies
the $b$-term. Confidence intervals are large-sample clustered intervals over
image-level units, not intervals requiring normally distributed individual errors.

\section{Experiments}
\label{sec:experiments}

FLIP is applied at inference time by replacing the post-normalization Final-site
tensor $\mathbf{z}$ with
$\widetilde{\mathbf{z}}=\max(\mathbf{z},\vartheta)$ prior to logit computation. Implementation details are provided in
Appendix~\ref{app:flip}; code, prompts, sweep grids, run manifests, and analysis
scripts are available in the github repository at
\url{https://github.com/earl-juanico/flip-qwen3vl/}.

\subsection{Experimental Setup}
Experiments use 7{,}056 clustered COCO image--query pairs, with localization and
counting queries issued per sample under matched intervention strengths. Because
multiple object queries may be drawn from the same image, statistical uncertainty
is computed at the image-cluster level rather than by treating image--query pairs
as independent. Reported 95\% confidence intervals therefore rely on the
large-sample approximation for the image-clustered estimator, not on normality of
individual errors or residuals. We use Qwen3-VL-4B-Instruct as the anchor model;
cross-architecture results are in Appendix~\ref{app:cross_architecture}. Unless otherwise noted, the core intervention sweeps use a fixed, pre-specified
$\vartheta$ grid and fixed random seeds for replicability; repeated-run dispersion
is reported only where runtime-level nondeterminism cannot be controlled in the
serving stack. We treat pointwise tests across $\vartheta$ as descriptive
screening rather than as a family of confirmatory discoveries: no claim depends on
declaring any single scanned $\vartheta$ significant. Regime identification is
based on the full sweep and the FWHM window around the primary $\dRfifty$ peak,
then checked against grounding-proxy alignment, feature-coherence controls,
negative controls, raw-layer site specificity, and image-cluster subsampling.
Hardware details for the core sweeps are reported in Appendix~\ref{app:flip}.

\paragraph{Evaluation Protocol (Behavioral Readout).}
All interventions are evaluated through final model outputs. For raw decoder-layer
checks, we apply $h^{(\ell)} \rightarrow \tilde h^{(\ell)}(\vartheta)$ at the
selected layer and run the remaining computation to logits and decoded
predictions. For the \emph{Final} site, we intervene on the post-normalization
state passed to the output head. Thus, all layer checks measure downstream
behavior, not intermediate-state change alone.

\subsection{Metrics}

\paragraph{Primary quantities.}
We use $\Rfifty$ / $\dRfifty$ as the detection-side mediator and $\ecount$ as
the downstream outcome.

\paragraph{Attribution redistribution (RMR).}
Relevance mass reallocation (RMR) is an appendix-side redistribution diagnostic rather than a causal measure.
We track $\mathrm{RMR}_{+}$, $\mathrm{RMR}_{-}$, and
$\mathrm{RMR}_{\mathrm{obj}}$; construction and interpretation are given in
Appendices~\ref{app:rmr_impl} and~\ref{app:rmr_interpretation}.

\begin{table*}[t]
\centering
\caption{\textbf{Delta-method grounding-proxy product summary on Qwen3-VL-4B.}
$a$: effect of FLIP on $\Delta R_{50}$; $b$: Poisson-GLM association of
$\Delta R_{50}$ with $\log \mathbb{E}[\mathcal{E}_{\mathrm{count}}]$ on strict
counting responses; $a\times b$: product summary; and
$\mathrm{IRR}_{\mathrm{indirect}}=\exp(a\times b)$. Values below $1$ indicate
that improved recall is associated with reduced tolerant counting error. The
quantity is interpreted as grounding-proxy compatibility, not direct within-pass
causal mediation.}
\label{tab:mediation-4b}
\scriptsize
\setlength{\tabcolsep}{4pt}
\begin{tabular}{lcccccccc}
\toprule
$\vartheta$ & $a$ & SE$(a)$ & $b$ & SE$(b)$ & $a\!\times\!b$ & SE$(a\!\times\!b)$ & $p$-value & $\mathrm{IRR}_{\mathrm{indirect}}$ [95\% CI] \\
\midrule
$-50.0$ & 0.00208 & 0.00054 & $-2.070$ & 0.158 & $-0.00430$ & 0.00117 & $0.0002$ &  0.9957 [0.9934, 0.9980] \\
$-5.0$ & 0.02677 & 0.00214 & $-2.092$ & 0.152 & $-0.05600$ & 0.00606 & $<0.0001$ &  0.9455 [0.9344, 0.9568] \\
$-4.0$ & 0.02962 & 0.00219 & $-2.087$ & 0.151 & $-0.06180$ & 0.00638 & $<0.0001$ & 0.9401 [0.9284, 0.9519] \\
$-2.0$ & 0.03878 & 0.00242 & $-2.079$ & 0.148 & $-0.08061$ & 0.00765 & $<0.0001$ & 0.9226 [0.9088, 0.9365] \\
$-1.0$ & 0.04832 & 0.00277 & $-2.049$ & 0.146 & $-0.09903$ & 0.00905 & $<0.0001$ & 0.9057 [0.8898, 0.9219] \\
$-0.5$ & 0.05283 & 0.00294 & $-2.015$ & 0.136 & $-0.10647$ & 0.00932 & $<0.0001$ & 0.8990 [0.8827, 0.9156] \\
$-0.3$ & 0.05777 & 0.00299 & $-2.032$ & 0.137 & $-0.11739$ & 0.01001 & $<0.0001$ & 0.8892 [0.8720, 0.9069] \\
$-0.2$ & 0.05782 & 0.00304 & $-2.027$ & 0.137 & $-0.11723$ & 0.01002 & $<0.0001$ &  0.8894 [0.8721, 0.9070] \\
$0.0$ & $0.05900$ & 0.00321 & $-2.074$ & 0.133 & $-0.12236$ & 0.01028 & $<0.0001$ & 0.8848 [0.8672, 0.9028] \\
$0.5$ & $0.02638$ & 0.00355 & $-2.151$ & 0.138 & $-0.05674$ & 0.00847 & $<0.0001$ & 0.9448 [0.9293, 0.9607] \\
$1.0$ & $-0.531480$ & 0.00590 & $-2.069$ & 0.157 & $1.09974$ & 0.08454 & $<0.0001$ & 3.0034 [2.5447, 3.5447] \\
\bottomrule
\end{tabular}
\end{table*}

\begin{figure}[t]
    \centering
    \includegraphics[
    scale=0.35,
    trim=8cm 2cm 4cm 1cm,
    clip
  ]{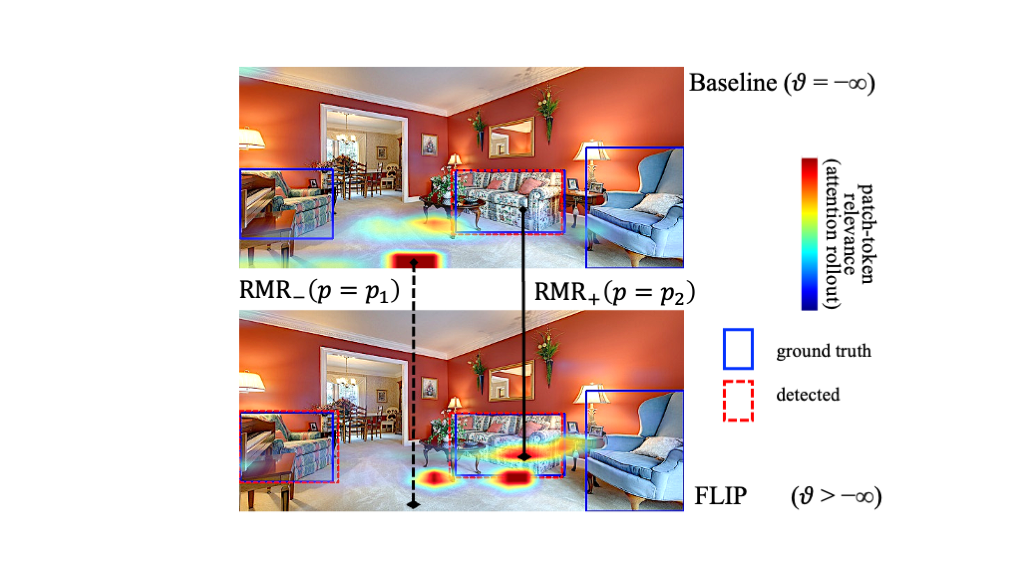}
\caption{\textbf{Attribution redistribution under FLIP for a single image--prompt pair.}
Top: baseline target-conditioned patch relevance. Bottom: relevance after
intervention. The figure qualitatively illustrates the bounded redistribution
associated with the interior regime; the saliency construction is described in
Appendix~\ref{app:rmr_impl}, and the aggregate RMR diagnostics are reported in
Appendix~\ref{app:rmr_dispersion}. Representative example only.}
\label{fig:rmr_example}
\end{figure}
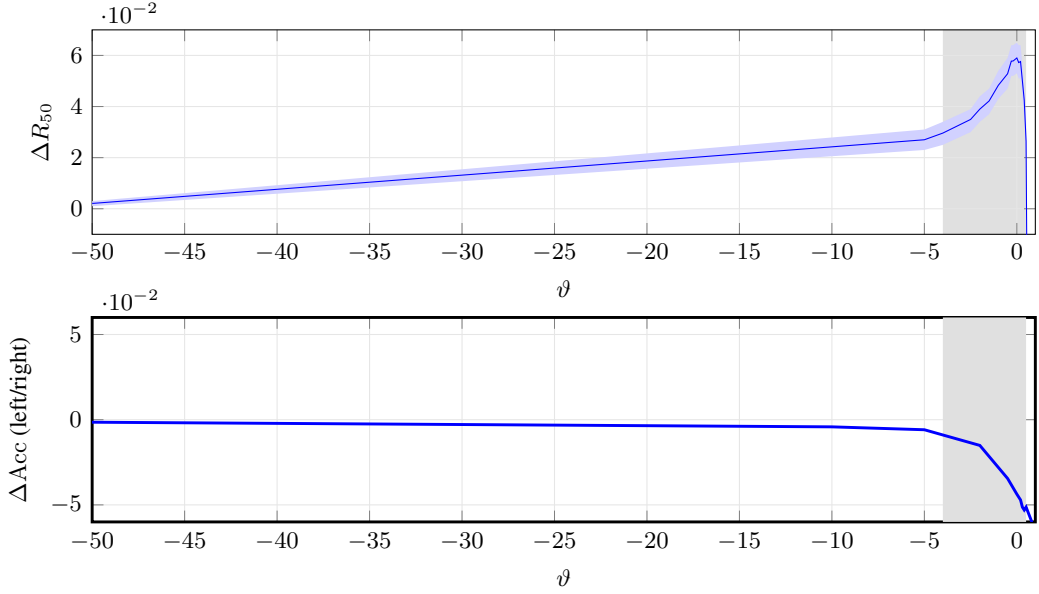
\begin{figure*}
\centering
\begin{tikzpicture}

\begin{axis}[
    name=toppanel,
    width=0.82\textwidth,
    height=0.25\textwidth,
    xlabel={$\vartheta$},
    ylabel={$\dRfifty$},
    xmin=-50, xmax=1.0,
    ymin=-0.01, ymax=0.07,
    tick label style={font=\small},
    label style={font=\small},
    grid=both,
    major grid style={draw=gray!20},
    minor grid style={draw=gray!10},
    line width=1.1pt,
]
\def\Ymax{0.07}
\def\Ymin{-0.01}
\def\Amin{-50}
\def\Amax{-4}
\def\Bmax{0.5}
\def\Cmax{1}
\addplot[draw=none, fill=black!12, forget plot, on layer=axis background] coordinates {(\Amax,\Ymin) (\Bmax,\Ymin) (\Bmax,\Ymax) (\Amax,\Ymax)} \closedcycle;

\addplot[name path=qwen4b_lo, draw=none, forget plot] coordinates {
    (-50, 0.001)
    (-5, 0.023000)
    (-4, 0.025000)
    (-2.5, 0.030000)
    (-2.0, 0.034000)
    (-1.5, 0.037000)
    (-1.0, 0.043000)
    (-0.5, 0.047000)
    (-0.3, 0.052000)
    (-0.2, 0.052000)
    (0.0, 0.053000)
    (0.1, 0.051000)
    (0.2, 0.051000)
    (0.4, 0.036000)
    (0.5, 0.019000)
    (1.0, -0.543000)
};
\addplot[name path=qwen4b_hi, draw=none, forget plot] coordinates {
    (-50, 0.003)
    (-5, 0.031000)
    (-4, 0.034000)
    (-2.5, 0.039000)
    (-2.0, 0.044000)
    (-1.5, 0.047000)
    (-1.0, 0.054000)
    (-0.5, 0.059000)
    (-0.3, 0.064000)
    (-0.2, 0.064000)
    (0.0, 0.065000)
    (0.1, 0.064000)
    (0.2, 0.064000)
    (0.4, 0.049000)
    (0.5, 0.033000)
    (1.0, -0.520000)
};
\addplot[fill=blue!18, draw=none, forget plot] fill between[of=qwen4b_lo and qwen4b_hi];
\addplot[blue, forget plot] coordinates {
    (-50, 0.002079231)
    (-5, 0.0270000000000000)
    (-4, 0.0296154969316435)
    (-2.5, 0.0349537497319266)
    (-2.0, 0.0390000000000000)
    (-1.5, 0.0421748456120951)
    (-1.0, 0.0483197857298569)
    (-0.5, 0.0528268844010360)
    (-0.3, 0.0577666466549469)
    (-0.2, 0.0578242891486406)
    (0.0, 0.0590000000000000)
    (0.1, 0.0572127308350086)
    (0.2, 0.0575911772372242)
    (0.4, 0.0424744359138014)
    (0.5, 0.0263772618529984)
    (1.0, -0.5314803772390620)
};
\end{axis}

\begin{axis}[
    name=bottompanel,
    at={($(toppanel.south)+(0,-1.1cm)$)},
    anchor=north,
    width=0.82\textwidth,
    height=0.25\textwidth,
    xlabel={$\vartheta$},
    ylabel={$\dAcc$ (left/right)},
    xmin=-50, xmax=1.0,
    ymin=-0.06, ymax=0.06,
    log basis y=10,
    tick label style={font=\small},
    label style={font=\small},
    grid=both,
    major grid style={draw=gray!20},
    minor grid style={draw=gray!10},
    line width=1.1pt,
    legend style={
        at={(0.5,-0.42)},
        anchor=north,
        legend columns=3,
        draw=none,
        fill=none,
        font=\small
    },
]
\def\Ymax{0.06}
\def\Ymin{-0.06}
\def\Amin{-50}
\def\Amax{-4}
\def\Bmax{0.5}
\def\Cmax{1}
\addplot[draw=none, fill=black!12, forget plot, on layer=axis background] coordinates {(\Amax,\Ymin) (\Bmax,\Ymin) (\Bmax,\Ymax) (\Amax,\Ymax)} \closedcycle;

\addplot[blue] coordinates {
    (-60, -0.0008)
    (-10, -0.0042)
    (-5, -0.0059)
    (-2, -0.0151)
    (-0.5, -0.0345)
    (0.0, -0.0437)
    (0.2, -0.0471)
    (0.3, -0.0513)
    (0.4, -0.053)
    (0.5, -0.0513)
    (0.8, -0.0597)
    (1.0, -0.0706)
};

\end{axis}

\end{tikzpicture}
\caption{\textbf{Primary dose--response and negative control.}
Top: FLIP on Qwen3-VL-4B-Instruct yields the expected three-region detection
response in $\dRfifty$: near-baseline behavior, bounded interior gain, and
collapse under strong flooring. Bottom: the singleton-pair left/right control,
measured by baseline-centered $\dAcc$, does not reproduce this positive interior
regime; it remains near baseline over most non-collapse strengths and degrades
under stronger intervention. The shaded band marks the FWHM window from the
primary detection peak; bands denote 95\% CI where available.}
\label{fig:dose-response-detection}
\end{figure*}

\subsection{Results}
\label{sec:results}

\paragraph{Primary regime evidence.}
\Cref{fig:dose-response-detection} shows a three-regime response in $\dRfifty$:
negligible change, a bounded interior gain, and collapse. The interior regime is
not declared from the single largest point in the sweep, and we do not interpret
unadjusted pointwise $p$-values over $\vartheta$ as separate discoveries. Instead,
the peak defines a descriptive FWHM window that must survive the full protocol:
grounding-proxy alignment, feature-coherence disruption, negative-control contrast,
logit-facing site specificity, and subsample stability. Appendix~\ref{app:bootstrap}
(\Cref{fig:bootstrap}) shows that the same interior window remains stable under
image-cluster subsampling.

\paragraph{Grounding-proxy alignment.}
The interior regime aligns with downstream behavior: intermediate flooring
increases $\dRfifty$ while reducing $\ecount$, and over-suppression reverses the
effect. \Cref{tab:mediation-4b} shows a non-zero product summary in the same
$\vartheta$ window where recall peaks. Because localization and counting are
separate forward passes, this supports a shared grounding-proxy interpretation,
not a claim that one output directly causes the other.

\paragraph{Logit-facing site specificity.}
\Cref{fig:cross_layer_check} shows that hidden-state flooring alone does not
induce the Final-site signature. Raw decoder-layer interventions remain near
baseline, collapse, or show only partial movement, while the post-normalization
Final site exhibits the full interior regime. To test whether this contrast is a
raw-threshold artifact, we also overlay percentile-matched sweeps for
Layers~8,~16, and~32. These matched raw-layer sweeps do not recover the
Final-site dose--response and instead attenuate or worsen the response. Thus,
the raw-layer failures are not explained by applying too weak or poorly scaled a
threshold. The full protocol---primary-task regime, left/right contrast,
grounding-proxy alignment, and feature-coherence dependence---localizes to the
post-normalization logit-facing state rather than to raw-layer depth or marginal
intervention strength alone.
\input{figure/main-cross-layer-check}

\begin{figure*}
\centering
\begin{tikzpicture}
  \begin{axis}[
    name=top,
    width=0.8\textwidth, height=3.5cm,
    xmin=-10, xmax=1, ymin=-0.1, ymax=0.1,
    xtick={-25,-20,-15,-10,-5,0,5},
    xticklabels={}, ytick={-0.1,0,0.1},
    grid=both, major grid style={draw=gray!30}, minor grid style={draw=gray!15},
    legend style={at={(0.02,0.98)}, anchor=north west, fill=white, draw=none, font=\small},
    axis line style={black}, tick style={black},
  ]
    \addplot[very thick] coordinates {
    (-50,0.002079231)(-5,0.027)(-4,0.029615497)
    (-2.5,0.03495375)(-2,0.039)(-1.5,0.042174846)(-1,0.048319786)(-0.5,0.052826884)
    (-0.3,0.057766647)(-0.2,0.057824289)(0,0.059)(0.1,0.057212731)
    (0.2,0.057591177)(0.4,0.042474436)(0.5,0.026377262)(1,-0.531480377)
    };
    \addplot[thick, dashed] coordinates {
    (-50,-0.022046971)(-5,0.011626532)(-4,0.018008087)
    (-2.5,0.024415207)(-2,0.027806851)(-1.5,0.032197256)(-1,0.039069749)(-0.5,0.047788281)
    (-0.3,0.049373501)(-0.2,0.050453073)(0,0.049286348)(0.1,0.040855208)
    (0.2,0.019731438)(0.4,-0.071604502)(0.5,-0.146387356)(1,-0.532121403)
    };
    \addplot[thick, dashdotted] coordinates {
    (-50,0.014359591)(-5,0.037975212)(-4,0.040100187)
    (-2.5,0.043839058)(-2,0.046339536)(-1.5,0.046085269)(-1,0.043385381)(-0.5,0.006697833)
    (-0.3,-0.02338556)(-0.2,-0.044260681)(0,-0.083646173)(0.1,-0.101468386)
    (0.2,-0.135171712)(0.4,-0.279588983)(0.5,-0.371970577)(1,-0.532121403)
    };
    \addplot[thick, dashdotdotted] coordinates {
    (-50,0.002030795)(-5,0.034530078)(-4,0.040404267)
    (-2.5,0.048149145)(-2,0.049479375)(-1.5,0.0541114)(-1,0.06026972)(-0.5,0.055424791)
    (-0.3,0.043090911)(-0.2,0.034978707)(0,0.001869541)(0.1,-0.027177572)
    (0.2,-0.09004874)(0.4,-0.323502676)(0.5,-0.470411262)(1,-0.532121403)
    };
    \addplot[thick, densely dotted] coordinates {
    (-50,0.022758138)(-5,0.055365354)(-4,0.058514655)
    (-2.5,0.063544906)(-2,0.06605575)(-1.5,0.06971064)(-1,0.070709545)(-0.5,-0.294699128)
    (-0.3,-0.521584753)(-0.2,-0.527780431)(0,-0.529043819)(0.1,-0.530478273)
    (0.2,-0.531655109)(0.4,-0.532121403)(0.5,-0.532121403)(1,-0.532121403)
    };
  \end{axis}
  \begin{axis}[
    name=bottom, at={(top.south west)}, anchor=north west, yshift=-0.4cm,
    width=0.8\textwidth, height=3.0cm,
    xmin=-10, xmax=1, ymin=-0.6, ymax=-0.15,
    xlabel={$\vartheta$}, ylabel={$\dRfifty$},
    ylabel style={xshift=32pt},
    xtick={-25,-20,-15,-10,-5,0,5}, ytick={-0.6,-0.4,-0.2},
    grid=both, major grid style={draw=gray!30}, minor grid style={draw=gray!15},
    legend style={
  at={(0.02,0.98)}, 
  anchor=north west, 
  fill=white, 
  draw=none, 
  font=\small,
  legend columns=2
},
  ]
    \addplot[very thick] coordinates {
    (-50,0.002079231)(-5,0.027)(-4,0.029615497)
    (-2.5,0.03495375)(-2,0.039)(-1.5,0.042174846)(-1,0.048319786)(-0.5,0.052826884)
    (-0.3,0.057766647)(-0.2,0.057824289)(0,0.059)(0.1,0.057212731)
    (0.2,0.057591177)(0.4,0.042474436)(0.5,0.026377262)(1,-0.531480377)
    };
    \addlegendentry{$\lambda=0.00$}
    \addplot[thick, dashed] coordinates {
    (-50,-0.022046971)(-5,0.011626532)(-4,0.018008087)
    (-2.5,0.024415207)(-2,0.027806851)(-1.5,0.032197256)(-1,0.039069749)(-0.5,0.047788281)
    (-0.3,0.049373501)(-0.2,0.050453073)(0,0.049286348)(0.1,0.040855208)
    (0.2,0.019731438)(0.4,-0.071604502)(0.5,-0.146387356)(1,-0.532121403)
    };
    \addlegendentry{$\lambda=0.10$}  
    \addplot[thick, dashdotted] coordinates {
    (-50,0.014359591)(-5,0.037975212)(-4,0.040100187)
    (-2.5,0.043839058)(-2,0.046339536)(-1.5,0.046085269)(-1,0.043385381)(-0.5,0.006697833)
    (-0.3,-0.02338556)(-0.2,-0.044260681)(0,-0.083646173)(0.1,-0.101468386)
    (0.2,-0.135171712)(0.4,-0.279588983)(0.5,-0.371970577)(1,-0.532121403)
    };
    \addlegendentry{$\lambda=0.20$}
    \addplot[thick, dashdotdotted] coordinates {
    (-50,0.002030795)(-5,0.034530078)(-4,0.040404267)
    (-2.5,0.048149145)(-2,0.049479375)(-1.5,0.0541114)(-1,0.06026972)(-0.5,0.055424791)
    (-0.3,0.043090911)(-0.2,0.034978707)(0,0.001869541)(0.1,-0.027177572)
    (0.2,-0.09004874)(0.4,-0.323502676)(0.5,-0.470411262)(1,-0.532121403)
    };
    \addlegendentry{$\lambda=0.25$}
    \addplot[thick, densely dotted] coordinates {
    (-50,0.022758138)(-5,0.055365354)(-4,0.058514655)
    (-2.5,0.063544906)(-2,0.06605575)(-1.5,0.06971064)(-1,0.070709545)(-0.5,-0.294699128)
    (-0.3,-0.521584753)(-0.2,-0.527780431)(0,-0.529043819)(0.1,-0.530478273)
    (0.2,-0.531655109)(0.4,-0.532121403)(0.5,-0.532121403)(1,-0.532121403)
    };
    \addlegendentry{$\lambda=0.33$}
  \end{axis}
\end{tikzpicture}
\caption{\textbf{Graduated permutation control via mixing parameter $\lambda$.}
Both panels show $\Delta R_{50}$ under feature-coherence mixing, with all curves
centered against the same unpermuted no-intervention baseline
$(\lambda=0,\vartheta=-\infty)$ rather than against $\lambda$-specific permuted
baselines. The upper panel zooms the near-zero/interior-response range, while the
lower panel zooms the collapse regime. Relative to the unpermuted case
$(\lambda=0)$, all permuted conditions lose the positive regime earlier and enter
collapse sooner, although the ordering among nonzero $\lambda$ values is not
strictly monotonic. This indicates that feature coherence is necessary for the
stability of the interior regime, and that the observed loss is not an artifact
of subtracting an already degraded permuted baseline.}
\label{fig:graduated_permutation_control}
\end{figure*}
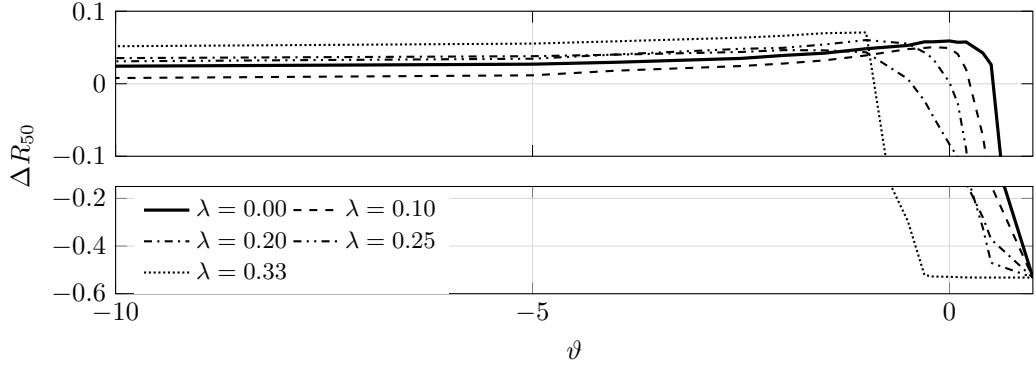

\paragraph{Specificity under controls.}
At the Final site, permutation controls
(\Cref{fig:graduated_permutation_control}) show that the positive interior
regime is not preserved once feature coherence is disrupted.
All $\dRfifty$ and mediator quantities in this control are centered against the
same unpermuted no-intervention baseline $(\lambda=0,\vartheta=-\infty)$, not
against a $\lambda$-specific permuted baseline. Thus any baseline damage from
permutation is retained rather than subtracted away. Relative to $\lambda=0$,
nonzero $\lambda$ settings lose the positive window and collapse earlier, though
their ordering is not strictly monotonic; the control is therefore interpreted as
a feature-coherence stress test rather than as an isolated flooring effect on an
already permuted model.

\paragraph{Operator comparison.}
Flooring, symmetric clipping, and additive shift each produce regime-shaped
primary-task curves, so non-monotonicity or peak $\dRfifty$ alone is not
specific evidence. The full operator replication
(Appendix~\ref{app:operator_baselines}, \Cref{fig:operator_comparison})
therefore evaluates the joint signature: grounding-proxy alignment,
feature-coherence dependence, logit-facing site specificity, and failure on the
left/right performance control.

\paragraph{Cross-architecture replication and broader spatial control.}
\Cref{fig:multimodel_composite_two_panel_centered} shows the same qualitative
three-phase structure in Qwen3-VL-8B, Qwen3-VL-4B, and Kimi-VL-A3B, with
Qwen3-VL-8B the lower-amplitude case. The bottom-panel singleton-pair left/right
control does not reproduce the coherent positive interior regime across models;
instead, it stays near baseline or drifts downward before late collapse; task
details are given in Appendix~\ref{app:leftright_control}. Appendix~\ref{app:cross_architecture}
adds the auxiliary indoor/outdoor response-switch diagnostic and mediation replications
(Tables~\ref{tab:mediation-kimi} and~\ref{tab:mediation-8b}).
\input{figure/main-cross-model-with-leftright-control}

\paragraph{Appendix-side corroboration.}
Appendix~\ref{app:rmr_interpretation} shows that, for the lower-amplitude
Qwen3-VL-8B case, $\mathrm{RMR}_{+}$ tracks the interior regime while
$\mathrm{RMR}_{-}$ and $\mathrm{RMR}_{\mathrm{obj}}$ mark the transition to
over-suppression. Appendix~\ref{app:benchmark_consistency} further shows
structured sweeps on MMStar, MindCube, and NaturalBench. These analyses broaden
the regime picture but remain secondary to the mediator-aligned protocol
evidence.

\section{Discussion}
\label{sec:discussion}

FLIP occupies the screening stage between broad behavioral probing and
circuit-level identification. Primary-task improvement is not enough, and neither
is proximity to the last decoder block. The relevant evidence is the joint
signature: bounded grounding-proxy-aligned response, Final-site
feature-coherence dependence, failure on a performance-based negative control,
and specificity to the normalized logit-facing readout. The raw-layer and
percentile-matched checks separate depth, dose, and validation: matching marginal
intervention strength at Layers~8,~16, and~32 does not recover the Final-site
response. Thus, FLIP does not claim that a circuit has been found; it nominates
the Final-site interior regime as the most justified target for patching, circuit
tracing, or causal-abstraction tests.

The controls define the scope of this interpretation. \Cref{tab:mediation-4b}
supports a shared grounding-proxy account in the interior regime and reversal
under over-suppression, while
\Cref{fig:multimodel_composite_two_panel_centered} shows cross-model persistence
despite effect-size variation. The left/right task rules out a broad
performance-gain account, and the indoor/outdoor diagnostic is retained only as
a baseline-relative instability check. FLIP is therefore a diagnostic triage
tool, not a deployment-safety certificate, optimization method, circuit
identification result, or proof of model understanding.

\paragraph{Limitations.}
FLIP validates intervention effects; it does not identify the circuit producing
them or close the gap from population-level triage to causal mechanism. The
grounding-proxy analysis is scoped to detection/counting, where $\dRfifty$ is
observable but not a literal within-pass mediator. Within this scope, we test
feature-coherence disruption, performance-based negative controls, same-site
operator replication, image-cluster subsampling, multiple VLMs, and raw-layer
site specificity. The raw-layer checks are conducted under the implemented
threshold sweep and do not replace percentile- or norm-matched cross-site
interventions. The raw-layer result is therefore best read as a readout-space
contrast under this sweep, not as a claim that task-relevant information is
absent from earlier layers. Bridging FLIP-positive sites to circuits through
matched sweeps, patching, causal abstraction, or circuit tracing remains future
work.

\section{Conclusion}
FLIP reframes final layer inference-time intervention as population-level effect
validation for mechanistic interpretability. Rather than treating a primary-task
gain as self-interpreting, the protocol requires grounding-proxy alignment,
feature-coherence dependence, performance-based negative-control contrast, and
specificity to the logit-facing readout. Raw-layer and percentile-matched checks
further rule out a simple threshold-calibration explanation. This separates
structured, task-linked sensitivity from generic perturbation and provides a
lightweight screen for deciding when downstream patching, circuit tracing, or
causal-abstraction analysis is warranted.

\section*{Acknowledgement}
  DE Juanico acknowledges the funding and technical support from the ERDT and the UP Diliman Artificial Intelligence Graduate Program.


\bibliographystyle{plainnat}

\clearpage
\appendix

\section{Generic Perturbation: Definition and Scope}
\label{app:generic_perturbation}

We use the term \emph{generic perturbation} to denote behavioral changes induced by
non-specific disruption of the representation, rather than alignment with
task-relevant intermediate computations.

Formally, let $h(\mathbf{z})$ denote a task-relevant readout and
$\widetilde{\mathbf{z}} = T(\mathbf{z})$ an intervention. We say that the induced
change $\Delta h = h(\widetilde{\mathbf{z}}) - h(\mathbf{z})$ reflects generic
perturbation if it is not systematically mediated by a task-relevant intermediate
quantity, such as localization recall in the detection/counting setting studied
here.

Under this definition, generic perturbation includes: (i) uniform magnitude
suppression, (ii) random feature misalignment, and (iii) global shifts that do
not preserve feature semantics. The distinction is operationalized in the main
text through grounding-proxy alignment, feature-coherence controls, and negative-control
tasks.

\section{Detection Matching and Mediator Construction}
\label{app:matching}

Given ground-truth boxes $\{g_i\}_{i=1}^{n_g}$ and predictions $\{d_j\}_{j=1}^{n_d}$:
IoU matrix $L$ with $L_{ij}=\mathrm{IoU}(g_i,d_j)$; class-consistency matrix
$N\in\{-1,+1\}$; confidence-weighted overlap $C_{ij}=s_j L_{ij}$. In our
implementation, $s_j$ is the normalized predicted-box area, used only to break ties
in matching. A one-to-one assignment under IoU threshold $0.50$ produces per-sample
true-positive counts $\TPfifty$, from which recall $\Rfifty$ is computed as in
\cref{sec:cca}. Mean IoU is used only for matching eligibility and not as
the primary mediator in the main analysis.

\section{Output-Head Alignment: Scope and Interpretation}
\label{app:operator_null}

The output-head alignment baseline used in the main text is a standard
pre-decoding calculation, not a new representation theory. Its purpose is to
make explicit why useful behavior under FLIP is non-trivial: the flooring
operator is not task-conditioned, so any beneficial effect must arise from how
the induced displacement projects onto task-relevant output-head directions.

For a final-token hidden state $\mathbf z_t$ and output matrix $W$, FLIP induces
\[
\widetilde{\ell}_t-\ell_t = W^\top\Delta_{\vartheta,t}.
\]
Thus the sign and size of any useful pre-decoding change depend on how the
flooring displacement projects onto output-head directions. For a log-sum-exp
margin between task-relevant tokens $Y^+$ and contrast tokens $Y^-$,
\[
m(\mathbf z_t)
=
\log\sum_{y\in Y^+}\exp(w_y^\top \mathbf z_t)
-
\log\sum_{y\in Y^-}\exp(w_y^\top \mathbf z_t),
\]
the first-order change is governed by
\[
\left\langle
\bar w_{Y^+}(\mathbf z_t)-\bar w_{Y^-}(\mathbf z_t),
\Delta_{\vartheta,t}
\right\rangle .
\]
A systematic improvement therefore requires positive alignment between the
non-targeted flooring displacement and task-relevant output-head directions.

This does not imply that $\Rfifty$ or $\ecount$ are continuous or monotone
functions of $\vartheta$. The empirical pipeline includes autoregressive
decoding, string parsing, bounding-box matching, IoU thresholding, and integer
count-error scoring. These operations can introduce discontinuities. Accordingly,
the paper does not claim a theorem about monotonic behavior of decoded metrics.

The role of the theory is instead to identify what a useful FLIP effect would
require at the logit-facing representation. Because FLIP does not encode the
object category, box coordinates, or count target, a stable interior gain is
unexpected under a generic perturbation account. The empirical protocol then tests whether the behavioral signature is consistent
with task-linked alignment: the gain should align with the detection-to-counting
mediator, weaken when feature/output-head coherence is disrupted, localize to the
normalized logit-facing site, and fail to reproduce on negative-control tasks.
\section{Grounding-Proxy Interpretation and Limitations}
\label{app:mediation_details}

The product summary
\[
\Iflip \rightarrow \dRfifty,\qquad
\ecount \sim \Iflip+\dRfifty
\]
is interpreted as grounding-proxy evidence, not proof of direct within-pass
mediation. Localization and counting are elicited by separate prompts, so the
observed localization output cannot literally cause the separately generated
counting output. Instead, $\dRfifty$ is used as an observable proxy for an
unobserved intervention-sensitive grounding state that may be expressed as both
improved localization and reduced counting error.

This analysis, therefore, supports the narrower claim that the interior regime is
compatible with a shared grounding factor. It does not identify the specific
circuit, exclude other unobserved mediators, or establish causal sufficiency.

A potential concern is that $\dRfifty$ and $\ecount$ may share a common
computational ancestor through predicted bounding boxes, making the mediation
relationship tautological. In the present design, localization and counting are
elicited by separate prompts and scored from independent model outputs, so the
observed association is not attributable to a shared output-level pipeline. The
remaining alternative is a common representational cause within the model, in
which the intervention jointly affects both localization and counting. This possibility is not fully excluded, but is constrained by the Final-site
permutation control and raw-layer checks, which localize the full signature to
the normalized logit-facing site rather than raw decoder-layer outputs.

\section{FLIP Deployment and Runtime Intervention}
\label{app:flip}

FLIP is implemented as an inference-time patch to logit computation: the
post-normalization Final-site tensor $\mathbf{z}\in\mathbb{R}^{n\times d}$ is
replaced by $\widetilde{\mathbf{z}}=\max(\mathbf{z},\vartheta)$ before
multiplication by the output projection $\mathbf{W}^{\top}$. This realizes the
flooring operator directly at the readout site.

\paragraph{Strength sweep.}
$\vartheta$ is provided externally (e.g., a config file), enabling deterministic
sweeps over a fixed grid without modifying parameters, tokenization, or decoding.

\paragraph{Compute resources.}
All COCO-based core sweeps, including the primary detection/counting runs,
negative controls, raw-layer checks, same-site operator comparisons,
feature-coherence controls, cross-architecture replications, and image-cluster
subsampling runs, were executed as single-GPU inference jobs on one
NVIDIA A100-SXM4-40GB GPU using NVIDIA-SMI/driver 570.195.03 and CUDA 12.8, with
GPU memory utilization limited to 0.9. The corresponding wall-clock/GPU-hour
requirements are reported in the supplemental run manifest; in our runs, a full
model-level core sweep required approximately \texttt{5-6} GPU-hours,
\texttt{52-53}  GPU-hours, and \texttt{189} GPU-hours
for Qwen3-VL-4B, Qwen3-VL-8B, and Kimi-VL-A3B, respectively. Post-hoc parsing,
scoring, GLM/mediation analysis, subsample aggregation, and figure generation were
CPU-side analyses and did not require GPU acceleration. Preliminary or failed runs
outside the reported experiments required approximately \texttt{8
GPU-hours}.

\paragraph{Why the Final site.}
Final-site interventions are minimally invasive: they leave early visual feature
extraction unchanged and perturb the normalized readout state where image/text
evidence is mapped to logits.

\section{Cross-Architecture Validation}
\label{app:cross_architecture}

\paragraph{Cross-architecture replication.}
\Cref{fig:multimodel_composite_two_panel_centered} shows cross-model
dose--response for Qwen3-VL-8B, Qwen3-VL-4B, and Kimi-VL-A3B. Kimi-VL-A3B uses a
different visual encoder, cross-modal fusion mechanism, and training recipe from
the Qwen family, providing genuine cross-architecture replication. All three
models exhibit the same qualitative three-phase structure. Importantly, the
absolute gains are not uniformly tiny: the inset makes clear that Qwen3-VL-8B is
the lower-amplitude case, while Qwen3-VL-4B and Kimi-VL-A3B show visibly larger
interior peaks.

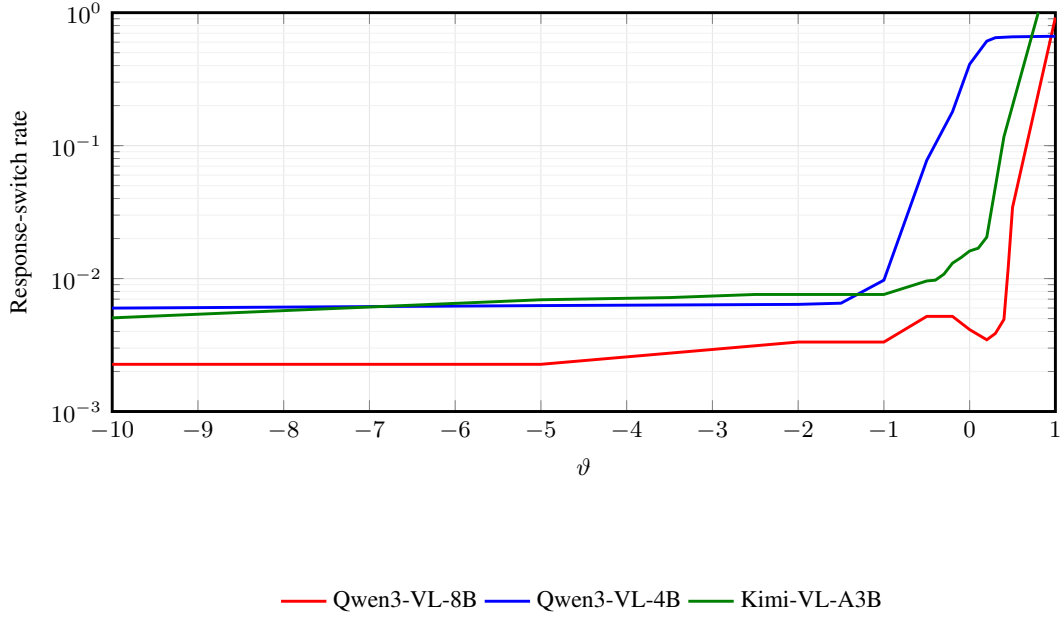
\begin{figure*}
\centering
\begin{tikzpicture}

\begin{axis}[
    name=bottompanel,
    at={($(toppanel.south)+(0,-2.2cm)$)},
    anchor=north,
    width=0.82\textwidth,
    height=0.4\textwidth,
    xlabel={$\vartheta$},
    ylabel={Response-switch rate},
    xmin=-10, xmax=1,
    ymode=log,
    ymin=0.001, ymax=1,
    log basis y=10,
    tick label style={font=\small},
    label style={font=\small},
    grid=both,
    major grid style={draw=gray!20},
    minor grid style={draw=gray!10},
    line width=1.1pt,
    legend style={
        at={(0.5,-0.42)},
        anchor=north,
        legend columns=3,
        draw=none,
        fill=none,
        font=\small
    },
]

\addplot[red] coordinates {
    (-10, 0.00226485478284039)
    (-5, 0.00226485478284039)
    (-2, 0.00333066879829469)
    (-1.5, 0.00333066879829469)
    (-1, 0.00333066879829469)
    (-0.5, 0.00519584332533972)
    (-0.2, 0.00519584332533972)
    (0.0, 0.00413002930988542)
    (0.2, 0.00346389555022648)
    (0.3, 0.00386357580602184)
    (0.4, 0.00492938982147615)
    (0.45, 0.0121236344257927)
    (0.5, 0.0342392752464694)
    (1.0, 0.919397815081268)
};
\addlegendentry{Qwen3-VL-8B}

\addplot[blue] coordinates {
    (-50, 0.004130029)
    (-10, 0.00599520383693045)
    (-5, 0.00626165734079403)
    (-2, 0.00639488409272581)
    (-1.5, 0.00652811084465760)
    (-1.0, 0.00972555289102051)
    (-0.5, 0.0775379696243005)
    (-0.2, 0.1798561151079130)
    (0.0, 0.4096722621902470)
    (0.2, 0.6109778843591790)
    (0.3, 0.6484146016520110)
    (0.5, 0.6582733812949640)
    (1.0, 0.6648014921396210)
};
\addlegendentry{Qwen3-VL-4B}

\addplot[green!50!black] coordinates {
    (-50.0, 0)
    (-30.0, 0)
    (-20.0, 0)
    (-10.0, 0.005062617)
    (-5.0, 0.006927791)
    (-3.5, 0.007194245)
    (-2.5, 0.007593925)
    (-2.0, 0.007593925)
    (-1.5, 0.007593925)
    (-1.0, 0.007593925)
    (-0.5, 0.009592326)
    (-0.4, 0.009725553)
    (-0.3, 0.010791367)
    (-0.2, 0.013056222)
    (-0.1, 0.014388489)
    (0.0, 0.016120437)
    (0.1, 0.016919797)
    (0.2, 0.02051692)
    (0.4, 0.116706635)
    (0.8, 0.9998667732480680) %
};
\addlegendentry{Kimi-VL-A3B}

\end{axis}

\end{tikzpicture}
\caption{\textbf{Indoor/outdoor response switch-rate comparison.}
Response switch-rate curves (defined in Appendix~\ref{app:sanity}) across Qwen3-VL-8B, Qwen3-VL-4B and Kimi-VL-A3B.}
\label{fig:indoor_outdoor_cross_model}
\end{figure*}

\paragraph{Cross-model instability diagnostic.}
The corresponding three-model indoor/outdoor response-switch diagnostic is shown
in \Cref{fig:indoor_outdoor_cross_model}. This diagnostic measures answer changes
relative to baseline rather than task performance. We therefore use it only to
check generic output instability; the performance-based negative-control contrast
is supplied by the left/right accuracy results in the main text and
Appendix~\ref{app:leftright_control}.

\paragraph{Cross-architecture mediation replications.}
To test whether the mediated detection-to-counting pathway extends beyond the
anchor model, we report the corresponding delta-method product summary tables for
Kimi-VL-A3B and Qwen3-VL-8B. These replications preserve the same qualitative
interpretation as the main-text Qwen3-VL-4B table: an interior regime with
$\mathrm{IRR}_{\mathrm{indirect}}<1$ and reversal under over-suppression.

\begin{table*}
\centering
\caption{\textbf{Delta-method grounding-proxy association for
$\Iflip$, $\dRfifty$, and $\ecount$ on Kimi-VL-A3B.}
$a$: effect of FLIP on the recall proxy $\dRfifty$; $b$: Poisson-GLM association
of $\dRfifty$ with $\log \mathbb{E}[\ecount]$; $a\!\times\!b$: product summary;
$\mathrm{IRR}_{\mathrm{indirect}}=\exp(a\!\times\!b)$, interpreted as a
grounding-proxy association rather than direct within-pass mediation. Values below
1 indicate that improved recall is associated with reduced tolerant counting
error.}
\label{tab:mediation-kimi}
\scriptsize
\setlength{\tabcolsep}{4pt}
\begin{tabular}{lcccccccc}
\toprule
$\vartheta$ & $a$ & SE$(a)$ & $b$ & SE$(b)$ & $a\!\times\!b$ & SE$(a\!\times\!b)$ & $p$-value & $\mathrm{IRR}_{\mathrm{indirect}}$ [95\% CI] \\
\midrule
$-3.5$ & 0.00419 & 0.00231 & $-2.659$ & 0.114 & $-0.01115$ & 0.00615 & $0.0697$ &  0.9889 [0.9771, 1.0009] \\
$-3.0$ & 0.00189 & 0.00241 & $-2.645 $ & 0.115 & $-0.0050$ & $0.0064$ & 0.4342 &  0.9950 [0.9826, 1.0075] \\
$-2.5$ & $-0.00020$ & 0.00257 & $-2.668$ & 0.110 & $0.00054$ & 0.00685 & $0.9376$ & 1.0005 [0.9872, 1.0141] \\
$-2.0$ & 0.00308 & 0.00269 & $-2.659$ & 0.113 & $-0.00819$ & 0.00715 & $0.2523$ & 0.9918 [0.9780, 1.0058] \\
$-1.5$ & 0.00841 & 0.00292 & $-2.657$ & 0.108 & $-0.02236$ & 0.00780 & $0.0042$ & 0.9779 [0.9630, 0.9930] \\
$-1.0$ & 0.02151 & 0.00312 & $-2.719$ & 0.100 & $-0.05847$ & 0.00875 & $<0.0001$ & 0.9432 [0.9272, 0.9595] \\
$-0.5$ & 0.03977 & 0.00357 & $-2.733$ & 0.095 & $-0.10866$ & 0.01047 & $<0.0001$ & 0.8970 [0.8788, 0.9156] \\
$-0.4$ & 0.04543 & 0.00368 & $-2.772$ & 0.092 & $-0.12595$ & 0.01102 & $<0.0001$ & 0.8817 [0.8628, 0.9009] \\
$-0.3$ & 0.04925 & 0.00374 & $-2.767$ & 0.091 & $-0.13624$ & 0.01129 & $<0.0001$ & 0.8726 [0.8535, 0.8922] \\
$-0.2$ & 0.05015 & 0.00382 & $-2.694$ & 0.103 & $-0.13510$ & 0.01153 & $<0.0001$ & 0.8736 [0.8541, 0.8936] \\
$-0.1$ & $0.04375$ & 0.00383 & $-2.635$ & 0.097 & $-0.11529$ & 0.01095 & $<0.0001$ & 0.8911 [0.8722, 0.9104] \\
$0.0$ & $0.03282$ & 0.00392 & $-2.624$ & 0.099 & $-0.08613$ & 0.01078 & $<0.0001$ & 0.9175 [0.8983, 0.9371] \\
$0.1$ & $0.01402$ & 0.00408 & $-2.632$ & 0.103 & $-0.03690$ & 0.01083 & $0.0007$ & 0.9638 [0.9435, 0.9844] \\
$0.2$ & $-0.01465$ & 0.00407 & $-2.564$ & 0.091 & $0.03755$ & 0.01051 & $0.0004$ & 1.0383 [1.0171, 1.0599] \\
$0.3$ & $-0.05500$ & 0.00395 & $-2.572$ & 0.094 & $0.14144$ & 0.01140 & $<0.0001$ & 1.1519 [1.1265, 1.1780] \\
$0.8$ & $-0.23269$ & 0.00535 & $-2.638$ & 0.127 & $0.61381$ & 0.03279 & $<0.0001$ & 1.8475 [1.7325, 1.9701] \\
$1.0$ & $-0.23269$ & 0.00535 & $-2.638$ & 0.127 & $0.61381$ & 0.03279 & $<0.0001$ & 1.8475 [1.7325, 1.9701] \\
\bottomrule
\end{tabular}
\end{table*}

\begin{table*}
\centering
\caption{\textbf{Delta-method grounding-proxy association for
$\Iflip$, $\dRfifty$, and $\ecount$ on Qwen3-VL-8B.}
$a$: effect of FLIP on the recall proxy $\dRfifty$; $b$: Poisson-GLM association
of $\dRfifty$ with $\log \mathbb{E}[\ecount]$; $a\!\times\!b$: product summary;
$\mathrm{IRR}_{\mathrm{indirect}}=\exp(a\!\times\!b)$, interpreted as a
grounding-proxy association rather than direct within-pass mediation. Values below
1 indicate that improved recall is associated with reduced tolerant counting
error.}
\label{tab:mediation-8b}
\scriptsize
\setlength{\tabcolsep}{4pt}
\begin{tabular}{lcccccccc}
\toprule
$\vartheta$ & $a$ & SE$(a)$ & $b$ & SE$(b)$ & $a\!\times\!b$ & SE$(a\!\times\!b)$ & $p$-value & $\mathrm{IRR}_{\mathrm{indirect}}$ [95\% CI] \\
\midrule
$-60.0$ & 0.00036 & 0.00024 & $-2.240$ & 0.121 & $ -0.00080$ & 0.00053 & $0.1319$ &  0.9992 [0.9982, 1.0002] \\
$-40.0$ & 0.00179 & 0.00054 & $-2.252$ & 0.119 & $-0.00402$ & 0.00124 & $0.0012$ &  0.9960 [0.9936, 0.9984] \\
$-20.0$ & 0.00532 & 0.00106 & $-2.296$ & 0.116 & $-0.01220$ & 0.00251 & $<0.001$ & 0.9879 [0.9830, 0.9927] \\
$-10.0$ & 0.00519 & 0.00128 & $-2.272$ & 0.119 & $-0.01180$ & 0.00297 & $<0.001$ & 0.9883 [0.9825, 0.9940] \\
$-5.0$ & 0.00693 & 0.00155 & $-2.272$ & 0.123 & $-0.01575$ & 0.00362 & $<0.001$ & 0.9844 [0.9774, 0.9914] \\
$-4.0$ & 0.00766 & 0.00160 & $-2.279$ & 0.115 & $-0.01747$ & 0.00375 & $<0.001$ & 0.9827 [0.9755, 0.9899] \\
$-2.5$ & 0.00800 & 0.00158 & $-2.248$ & 0.118 & $-0.01799$ & 0.00368 & $<0.001$ & 0.9822 [0.9751, 0.9893] \\
$-2.0$ & 0.00592 & 0.00153 & $-2.254$ & 0.112 & $-0.01335$ & 0.00351 & $<0.001$ & 0.9867 [0.9800, 0.9935] \\
$-1.5$ & 0.00394 & 0.00153 & $-2.276$ & 0.115 & $-0.00897$ & 0.00350 & $0.010$ & 0.9911 [0.9843, 0.9979] \\
$-1.0$ & 0.00327 & 0.00157 & $-2.249$ & 0.119 & $-0.00735$ & 0.00355 & $0.038$ & 0.9927 [0.9858, 0.9996] \\
$-0.5$ & $-0.00232$ & 0.00188 & $-2.222$ & 0.113 & $0.00515$ & 0.00418 & $0.218$ & 1.0052 [0.9970, 1.0134] \\
$0.0$ & $-0.01799$ & 0.00255 & $-2.312$ & 0.110 & $0.04160$ & 0.00622 & $<0.001$ & 1.0425 [1.0299, 1.0552] \\
\bottomrule
\end{tabular}
\end{table*}
\section{Raw Decoder-Layer Checks and Logit-Facing Specificity}
\label{app:cross_layer_failure}

This appendix reports raw decoder-layer checks under the same behavioral readout
used in the main text. The goal is to test whether the full protocol is
selectively satisfied at the normalized logit-facing Final site or appears
trivially in raw decoder-layer spaces. All checks are evaluated through final
outputs, so any effect must propagate to decoded behavior. The full result is
shown in \Cref{fig:cross_layer_failure_full}.

These checks separate depth, dose, and validation. Early/middle layers remain
near baseline before collapse, while late raw layers may show partial
$\dRfifty$ movement. We additionally include percentile-matched raw-layer sweeps
for Layers~8,~16, and~32 to test whether the failure is explained by marginal
threshold mismatch. These matched sweeps do not recover the Final-site response
and instead attenuate or worsen the raw-layer dose--response. The accepted
protocol, however, is not defined by nonzero $\dRfifty$ alone: it requires
negative-control contrast, grounding-proxy alignment, and feature-coherence
dependence. The post-normalization Final curve is the logit-facing reference;
the raw last-block output is a separate pre-normalization site.

\paragraph{Active-set and magnitude calibration.}
For each site, we track token-level active-set fraction
$A_{\ell,t}(\vartheta)$ and mean clamp magnitude $M_{\ell,t}(\vartheta)$
(\Cref{sec:A_and_M,tab:am_calibration}). $A$ measures how many hidden dimensions
are touched; $M$ measures average displacement. Layers~8 and~16 are not inactive:
at $\vartheta=0$, their $A$ values match the Final site, but their $M$ values are
roughly an order of magnitude smaller, followed by rapid saturation. Layer~32 is
closer in $A/M$ and can move $\dRfifty$, but still lacks the full
feature-coherence-dependent signature. Thus raw-layer failures are not explained
by too few clamped dimensions; the complete protocol localizes to the
post-normalization logit-facing site. The percentile-matched overlays in \Cref{fig:cross_layer_check} further show
that this localization is not removed by matching the marginal intervention
scale at representative early, middle, and late raw-layer sites.

\begin{table}[H]
\centering
\scriptsize
\setlength{\tabcolsep}{3.5pt}
\caption{\textbf{Token-level calibration for raw-layer checks.}
Means of active-set fraction $A$ and mean clamp magnitude $M$ under unpermuted
flooring. Early/middle layers show matched $A$ but much smaller $M$ at
$\vartheta=0$, then rapid saturation; Layer~32 is closer to the Final site but
does not reproduce the full Final-site signature.}
\label{tab:am_calibration}
\begin{tabular}{crrrrr}
\toprule
Layer & $\vartheta$ & Final $A$ & Final $M$ & Layer $A$ & Layer $M$ \\
\midrule
8  & 0.0 & 0.500 & 0.968 & 0.500 & 0.091 \\
8  & 0.5 & 0.602 & 1.122 & 0.979 & 0.504 \\
8  & 1.0 & 0.662 & 1.453 & 0.999 & 1.001 \\
\midrule
16 & 0.0 & 0.498 & 0.963 & 0.497 & 0.102 \\
16 & 0.5 & 0.591 & 1.039 & 0.980 & 0.508 \\
16 & 1.0 & 0.631 & 1.317 & 0.998 & 1.006 \\
\midrule
32 & 0.0 & 0.454 & 0.821 & 0.498 & 0.639 \\
32 & 0.5 & 0.535 & 1.000 & 0.643 & 0.926 \\
32 & 1.0 & 0.608 & 1.192 & 0.758 & 1.277 \\
\bottomrule
\end{tabular}
\end{table}

\input{figure/cross-layer-flip}
\FloatBarrier

\section{Redistribution Breadth and Dispersion Diagnostics}
\label{app:rmr_dispersion}

\subsection{Target-Conditioned Patch Relevance and RMR Computation}
\label{app:rmr_impl}

This appendix describes how target-conditioned patch relevance maps are computed
and how redistribution diagnostics are derived from them. These quantities are used
only as comparative diagnostics of redistribution geometry: they do not replace the
primary mediator analysis based on $\Delta R_{50}$ and
$\mathcal{E}_{\mathrm{count}}$.

\paragraph{Target-conditioned saliency maps.}
For each image--prompt pair, we compute a baseline patch relevance map and a
FLIP-intervened patch relevance map conditioned on a specified target token
sequence, such as the first \texttt{bbox\_2d} field or a generated detection
answer. The relevance pipeline first identifies the target token positions in the
full sequence (prompt plus generated response). We then define
$y_{\mathrm{target}}$ as a scalar target score for that sequence, computed as the
sum of the logits assigned to the identified target tokens, and compute
layerwise gradient-weighted attention relevance maps of the form
\begin{equation}
    R^{(\ell)} =
    \mathrm{ReLU}\!\left(
    \frac{\partial y_{\mathrm{target}}}{\partial A^{(\ell)}} \odot A^{(\ell)}
    \right),
\end{equation}
averaged over heads. These per-layer maps are combined into a joint relevance
matrix by augmenting with the identity, row-normalizing, and multiplying
cumulatively across layers.

To sharpen target specificity, the implementation also computes token-level
Grad\,$\times$\,Hidden importance from the final hidden state,
\begin{equation}
    g_i =
    \left\langle
    h_i,\,
    \frac{\partial y_{\mathrm{target}}}{\partial h_i}
    \right\rangle,
\end{equation}
with non-target positions masked to zero. The masked token-importance vector is
propagated through the joint relevance matrix and restricted to the visual
patch-token indices, yielding one scalar relevance value per image patch. For
Qwen3-VL, these values are arranged on the model's $24\times24$ patch grid.
Qualitative heatmaps are produced by interpolation and overlay, but all reported
statistics are computed directly from the underlying patch relevance vectors.

\paragraph{Baseline and FLIP conditions.}
The same target-conditioned relevance procedure is run twice: once under the
baseline condition and once under FLIP. Let
$s_{-\infty}\in\mathbb{R}^{P}$ denote the baseline patch saliency map and
$s_{\vartheta}\in\mathbb{R}^{P}$ the FLIP-intervened map, where $P$ is the number
of image patches. We define the patchwise saliency shift
\begin{equation}
    \delta = s_{\vartheta} - s_{-\infty}.
\end{equation}
All redistribution diagnostics are computed from $\delta$.

\paragraph{Directional RMR diagnostics.}
Let $\mathcal O$ denote the set of object patches under the patch mask induced by
the parsed bounding box. We report three directional redistribution statistics:
\begin{align}
\mathrm{RMR}_{+}(\vartheta) &=
\sum_{p\in\mathcal O} \max(\delta_p,0),\\
\mathrm{RMR}_{-}(\vartheta) &=
\sum_{p\in\mathcal O} \max(-\delta_p,0),\\
\mathrm{RMR}_{\mathrm{obj}}(\vartheta) &=
\sum_{p\in\mathcal O} \delta_p
-
\sum_{p\notin\mathcal O} \delta_p.
\end{align}
Thus, $\mathrm{RMR}_{+}$ measures saliency mass shifted toward object patches,
$\mathrm{RMR}_{-}$ measures saliency mass shifted away from object patches
(reported as a positive magnitude), and
$\mathrm{RMR}_{\mathrm{obj}}$ measures the resulting object-versus-background
contrast. The object mask is derived from parsed bounding boxes projected onto the
patch grid, so $\mathrm{RMR}_{\mathrm{obj}}$ is a relative contrast measure rather
than a raw object saliency total.

\paragraph{Interpretive role.}
These quantities are used to refine the interpretation of the FLIP sweep. In the
negligible regime, directional redistribution remains weak. In the interior regime,
bounded positive redistribution toward object patches becomes detectable. Under
over-suppression, object-aligned gains weaken while off-object loss and
redistribution breadth increase. The diagnostics therefore complement the primary
behavioral analysis without being treated as formal mediators or faithful causal
explanations.

\subsection{Comparative role of rollout}
\label{app:rmr_rollout}

The target-conditioned saliency maps above are constructed using
gradient-weighted attention relevance propagation with rollout. Their most reliable
use in this paper is comparative: differences across $\vartheta$ for the same model,
prompt, and target are informative, whereas absolute saliency magnitudes should not
be over-interpreted. For coordinate-token targets, tokenization positions can shift
across interventions, so target indices are anchored to baseline outputs and reused
across FLIP runs. If a target token is absent at a given $\vartheta$, that sample is
marked missing for RMR computation at that $\vartheta$.

\subsection{Regime interpretation of the RMR diagnostics}
\label{app:rmr_interpretation}

The RMR statistics are used to relate redistribution geometry to the regime
structure observed in the main dose--response analysis. They are not formal
mediators and do not replace the primary evidence based on $\Delta R_{50}$ and
$\mathcal{E}_{\mathrm{count}}$. Their role is interpretive: to indicate how
target-conditioned relevance mass is reallocated across the FLIP sweep.

\paragraph{Negligible regime.}
At very negative $\vartheta$, both $\Delta R_{50}$ and $\mathrm{RMR}_{+}$ are
small, indicating that FLIP is effectively inactive at the redistribution level.
This is the redistribution-side signature of the negligible-change regime.
Consistently, $\mathrm{RMR}_{-}$ remains comparatively modest and
$\mathrm{RMR}_{\mathrm{obj}}$ stays near zero in this region.

\begin{figure*}[t]
\centering
\begin{tikzpicture}

\def\panelw{0.42\textwidth}
\def\panelh{0.27\textwidth}
\def\hsep{1.2cm}
\def\vsep{1.1cm}

\begin{axis}[
    at={(0,0)},
    anchor=north west,
    width=\panelw,
    height=\panelh,
    xlabel={$\vartheta$},
    ylabel={$\mathrm{RMR}_{+}$},
    title={(a)},
    xmin=-60, xmax=1,
    ymin=0, ymax=7,
    xtick={-60,-40,-20,-10,-5,0},
    grid=both,
    tick label style={font=\scriptsize},
    label style={font=\scriptsize},
    title style={font=\scriptsize},
]
\addplot[
    very thick,
    mark=*,
    mark size=1.6pt
] coordinates {
    (-60, 3.136374596)
    (-40, 3.63803893)
    (-20, 5.13518229)
    (-10, 5.835190134)
    (-5, 5.808029322)
    (-2.5, 5.591355633)
    (-2, 5.37549227)
    (-1.5, 5.112595298)
    (-1, 4.843288386)
    (0, 4.613055149)
    (0.5, 3.436008893)
    (1, 2.98800414)
};
\end{axis}

\begin{axis}[
    at={(\panelw+\hsep,0)},
    anchor=north west,
    width=\panelw,
    height=\panelh,
    xlabel={$\vartheta$},
    ylabel={$\mathrm{RMR}_{+}$},
    title={(b)},
    xmin=-60, xmax=1,
    xtick={-60,-40,-20,-10,-5,0},
    ymin=0, ymax=7,
    axis lines=box,
    axis line style={line width=0.6pt},
    grid=both,
    tick label style={font=\scriptsize},
    label style={font=\scriptsize},
    title style={font=\scriptsize},
    clip=false,
]
\addplot[
    very thick,
    mark=*,
    mark size=1.6pt
] coordinates {
    (-60, 3.136374596)
    (-40, 3.63803893)
    (-20, 5.13518229)
    (-10, 5.835190134)
    (-5, 5.808029322)
    (-2.5, 5.591355633)
    (-2, 5.37549227)
    (-1.5, 5.112595298)
    (-1, 4.843288386)
    (0, 4.613055149)
    (0.5, 3.436008893)
    (1, 2.98800414)
};
\end{axis}

\begin{axis}[
    at={(\panelw+\hsep,0)},
    anchor=north west,
    width=\panelw,
    height=\panelh,
    xmin=-60, xmax=1,
    ymin=-0.006, ymax=0.01,
    axis x line=none,
    axis y line=right,
    ylabel={$\Delta R_{50}$},
    xtick=\empty,
    tick label style={font=\scriptsize},
    label style={font=\scriptsize},
    clip=false,
]
\addplot[
    red, very thick, dotted,
    mark size=1.5pt
] coordinates {
    (-60, 0.000355271)
    (-40, 0.001786556)
    (-20, 0.005315107)
    (-15, 0.004731481)
    (-10, 0.005194146)
    (-8, 0.005458730)
    (-6, 0.006102237)
    (-5, 0.006929527)
    (-4, 0.007662581)
    (-3.5, 0.007656283)
    (-3, 0.007176356)
    (-2.5, 0.008004802)
    (-2.4, 0.007272605)
    (-2.3, 0.007108362)
    (-2.2, 0.005318894)
    (-2.1, 0.007716428)
    (-2.0, 0.005924498)
    (-1.9, 0.006253556)
    (-1.8, 0.005890915)
    (-1.7, 0.006801346)
    (-1.6, 0.005475282)
    (-1.5, 0.003941299)
    (-1.4, 0.005629722)
    (-1.3, 0.005248077)
    (-1.2, 0.004064429)
    (-1.0, 0.003268797)
    (-0.5, -0.002318907)
};
\end{axis}

\begin{axis}[
    at={(0,-\panelh-\vsep)},
    anchor=north west,
    width=\panelw,
    height=\panelh,
    xlabel={$\vartheta$},
    ylabel={$\mathrm{RMR}_{-}$},
    title={(c)},
    xmin=-60, xmax=1,
    xtick={-60,-40,-20,-10,-5,0},
    grid=both,
    tick label style={font=\scriptsize},
    label style={font=\scriptsize},
    title style={font=\scriptsize},
]
\addplot[
    very thick,
    mark=*,
    mark size=1.6pt
] coordinates {
    (-60, 2.802913055)
    (-40, 3.368222961)
    (-20, 5.194107492)
    (-10, 5.891125404)
    (-5, 6.282232538)
    (-2.5, 6.915489814)
    (-2, 7.358970661)
    (-1.5, 7.891515181)
    (-1, 8.537328013)
    (0, 10.81296052)
    (0.5, 14.86049677)
    (1, 15.3136646)
};
\end{axis}

\begin{axis}[
    at={(\panelw+\hsep,-\panelh-\vsep)},
    anchor=north west,
    width=\panelw,
    height=\panelh,
    xlabel={$\vartheta$},
    ylabel={$\mathrm{RMR}_{\mathrm{obj}}$},
    title={(d)},
    xmin=-60, xmax=1,
    xtick={-60,-40,-20,-10,-5,0},
    grid=both,
    tick label style={font=\scriptsize},
    label style={font=\scriptsize},
    title style={font=\scriptsize},
]
\addplot[
    very thick,
    mark=*,
    mark size=1.6pt
] coordinates {
    (-60, -0.200502622)
    (-40, -0.102667747)
    (-20, 0.14290788)
    (-10, 0.104537095)
    (-5, 0.335582311)
    (-2.5, 0.756517531)
    (-2, 1.166694893)
    (-1.5, 1.634072595)
    (-1, 2.198232021)
    (0, 3.43787754)
    (0.5, 6.987757699)
    (1, 7.01802075)
};
\end{axis}

\end{tikzpicture}
\caption{\textbf{Qwen3-VL-8B redistribution diagnostics across the FLIP sweep.}
(a) $\vartheta$ versus $\mathrm{RMR}_{+}$, which peaks in the same broad interior
region where the detection-side response is positive. (b) Dual-axis overlay of
$\vartheta$ versus $\mathrm{RMR}_{+}$ (left axis) and $\vartheta$ versus
$\Delta R_{50}$ (right axis), showing redistribution-side co-variation with the
behavioral dose--response. (c) $\vartheta$ versus $\mathrm{RMR}_{-}$, which grows
toward collapse and marks increasing off-object saliency loss. (d) $\vartheta$
versus $\mathrm{RMR}_{\mathrm{obj}}$, which remains modest through the interior
regime but spikes late, consistent with suppression-dominant contrast enhancement
rather than constructive redistribution. Together, the panels link the appendix RMR
diagnostics to the three-regime structure of the main FLIP sweep.}
\label{fig:rmr_four_panel}
\end{figure*}
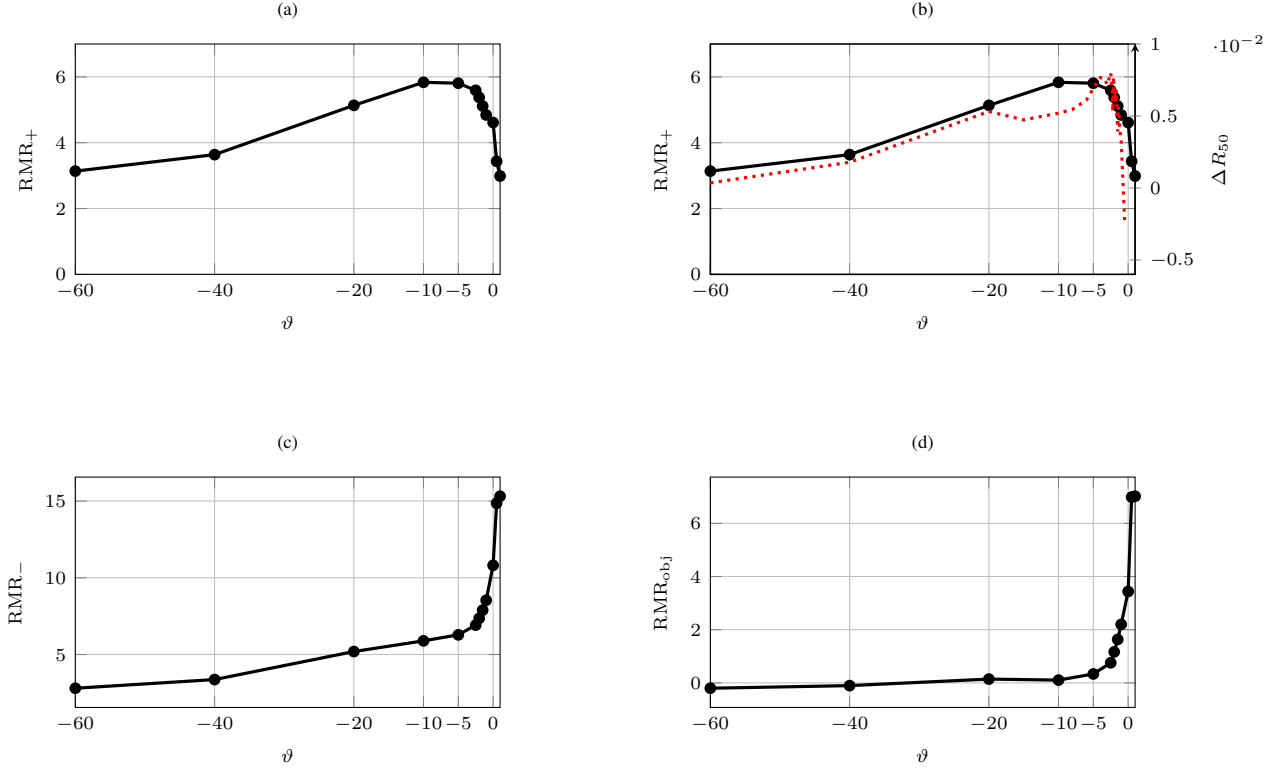

\paragraph{Interior regime.}
In the bounded interior regime, $\mathrm{RMR}_{+}$ rises and co-varies with
$\Delta R_{50}$, as shown directly in Figure~\ref{fig:rmr_four_panel}(b). This
indicates that the positive detection-side response is accompanied by bounded
object-aligned redistribution of relevance mass. In this regime,
$\mathrm{RMR}_{-}$ increases only gradually, and
$\mathrm{RMR}_{\mathrm{obj}}$ remains comparatively modest. The resulting picture is
one of constructive redistribution rather than generic suppression.

\paragraph{Over-suppression regime.}
As $\vartheta$ approaches the collapse region, $\mathrm{RMR}_{+}$ declines,
$\mathrm{RMR}_{-}$ rises sharply, and $\mathrm{RMR}_{\mathrm{obj}}$ spikes late.
This transition indicates that the regime has changed qualitatively. The late rise
in $\mathrm{RMR}_{\mathrm{obj}}$ should not be read as evidence of a stronger
constructive interior regime; instead, it is more consistent with
suppression-dominant contrast enhancement, in which object-relative saliency rises
because off-object mass is removed broadly.

\paragraph{Interpretive role.}
Taken together, the four-panel analysis refines the regime picture as follows:
the negligible regime shows little redistribution; the interior regime is the point
of strongest object-aligned reallocation, indexed by co-movement of
$\mathrm{RMR}_{+}$ and $\Delta R_{50}$; and the over-suppression regime is marked
by rising $\mathrm{RMR}_{-}$ and late spikes in
$\mathrm{RMR}_{\mathrm{obj}}$. Thus, the RMR diagnostics provide a redistribution-side
corroboration of the main regime structure without elevating curve shape alone into
a grounding-specificity claim.

To connect redistribution geometry more directly to the regime structure, we plot
$\mathrm{RMR}_{+}$, $\mathrm{RMR}_{-}$, and $\mathrm{RMR}_{\mathrm{obj}}$ against
$\vartheta$ for Qwen3-VL-8B, together with a direct overlay of
$\mathrm{RMR}_{+}$ and $\Delta R_{50}$.

\section{Vision-Centric Benchmark Consistency}
\label{app:benchmark_consistency}

This appendix reports additional results on vision-centric benchmarks to assess
whether the regime structure observed in the main text extends beyond the
detection-based evaluation setting.

\paragraph{Scope and role.}
These benchmarks are not designed to isolate the detection-derived mediator used
in the main analysis. As such, they do not constitute mediator-aligned evidence.
Instead, they serve as \emph{consistency checks} on whether intervention sweeps
produce structured, non-monotonic response patterns in broader evaluation
settings.

\begin{figure*}[t]
\centering
\begin{tikzpicture}

\pgfplotsset{
  left_axis/.style={
    width=0.65\textwidth,
    height=5.3cm,
    xmin=-50, xmax=5.0,
    xtick={-50,-40,-30,-20,-10,0},
    grid=both,
    major grid style={line width=0.2pt, draw=gray!25},
    minor grid style={line width=0.1pt, draw=gray!15},
    ticklabel style={font=\small},
    label style={font=\small},
    xlabel style={yshift=0pt},
    clip=true,
    ymin=-0.05, ymax=0.05,
    ylabel={percent accuracy change},
    xlabel={intervention strength $\vartheta$},
  },
    right_axis/.style={
      width=0.3\columnwidth,
      height=5.3cm,
      axis lines=none,
      xtick=\empty,
      ytick=\empty,
      xticklabels=\empty,
      yticklabels=\empty,
      legend style={
        font=\small, 
        draw=none, 
        fill=none,
        at={(0.1,0.9)}, 
        anchor=north west,
        legend cell align=left
      },
    }
}

\begin{groupplot}[
  group style={group size=2 by 1, horizontal sep=0.3cm}
]

\nextgroupplot[left_axis]

\addplot[draw=none, fill=black!12, forget plot] coordinates {(-4,-0.06) (0.5,-0.06) (0.5,0.07) (-4,0.07)} \closedcycle;

\addplot+[draw=blue, thick, dashed, mark=*,
  mark options={draw=blue, fill=blue}, error bars/.cd, y dir=both, y explicit,
] coordinates {
  (-60, 0.0021) +- (0, 0.0049)
  (-50, 0.0065) +- (0, 0.0028)
  (-40, 0.0109) +- (0, 0.0059)
  (-20, 0.0274) +- (0, 0.0054)
  (-10, 0.0334) +- (0, 0.0062)
  (-5, 0.0356) +- (0, 0.0045)
  (-3, 0.0347) +- (0, 0.0045)
  (-2.5, 0.0375) +- (0, 0.0048)
  (-2, 0.0388) +- (0, 0.0039)
  (-1.5, 0.0381) +- (0, 0.0032)
  (-1, 0.0360) +- (0, 0.0027)
  (-0.5, 0.0407) +- (0, 0.0058)
  (0, 0.0417) +- (0, 0.0076)
  (0.5, 0.0387) +- (0,0.0137)
  (1, -0.0351) +- (0,0.0078)
};

\addplot+[draw=black, thick, mark=square*,
  mark options={draw=black, fill=black}, error bars/.cd, y dir=both, y explicit,
] coordinates {
  (-60, 0.0095) +- (0, 0.0120)
  (-10, 0.0039) +- (0, 0.0026)
  (-4, 0.0101) +- (0, 0.0079)
  (-3.5, 0.0113) +- (0, 0.0136)
  (-3, 0.0197) +- (0, 0.0088)
  (-2.5, 0.0318) +- (0, 0.0098)
  (-2.2, 0.0324) +- (0, 0.014)
  (-2.12, 0.0315) +- (0, 0.0098)
  (-2.11, 0.0336) +- (0, 0.0109)  
  (-2.1, 0.0323) +- (0, 0.0117)
  (-2.0, 0.0343) +- (0, 0.0092)
  (-1.9, 0.0314) +- (0, 0.0043)
  (-1.8, 0.0343) +- (0, 0.010)
  (-1.5, 0.0242) +- (0, 0.0099)
  (-1, 0.0198) +- (0, 0.0088)
  (-0.5, -0.003) +- (0, 0.0086)
  (0, -0.0045) +- (0, 0.0311)
};

\addplot+[draw=red, thick, dashdotted, mark=diamond*,
  mark options={draw=red, fill=red}, error bars/.cd, y dir=both, y explicit,
] coordinates {
  (-50, -0.0043) +- (0,0)
  (-10, 0.0008) +- (0,0.0031)
  (-5, 0.0014)
  (-3, 0.0044) +- (0,0.0099)
  (-2.5, 0.0137) +- (0,0.001)
  (-2, 0.0100)  +- (0,0.0061)
  (-1.5, 0.0065) +- (0,0.005) 
  (-1, 0.0043) 
  (0.0, 0.0057) 
  (0.5, -0.0613)
  (1,-0.7728)
};

\addplot[black, thin, densely dotted, forget plot] coordinates {(-10,0) (1,0)};

\nextgroupplot[right_axis]

\addplot[draw=none, mark=none, forget plot] coordinates {(0,0)};  

\addlegendimage{blue, thick, dashed, mark=*}
\addlegendentry{MMStar}
\addlegendimage{black, thick, mark=square*}
\addlegendentry{MindCube}
\addlegendimage{red, thick, dashdotted, mark=diamond*}
\addlegendentry{NaturalBench}

\end{groupplot}
\end{tikzpicture}

\caption{\textbf{Dose--response of FLIP on vision-centric benchmarks.}
Curves show benchmark-specific performance changes relative to baseline. Error
bars denote standard deviation across repeated LMMS-Eval runs using independent
vLLM initializations of the same Qwen3-VL-8B model. The shaded band indicate the
regime partition defined from the primary detection/counting sweep under the
current $R_{50}$ mediator. MMStar, MindCube, and NaturalBench show structured, non-monotonic responses
consistent with the main regime picture.}
\label{fig:bench-dose-response-updated}
\end{figure*}
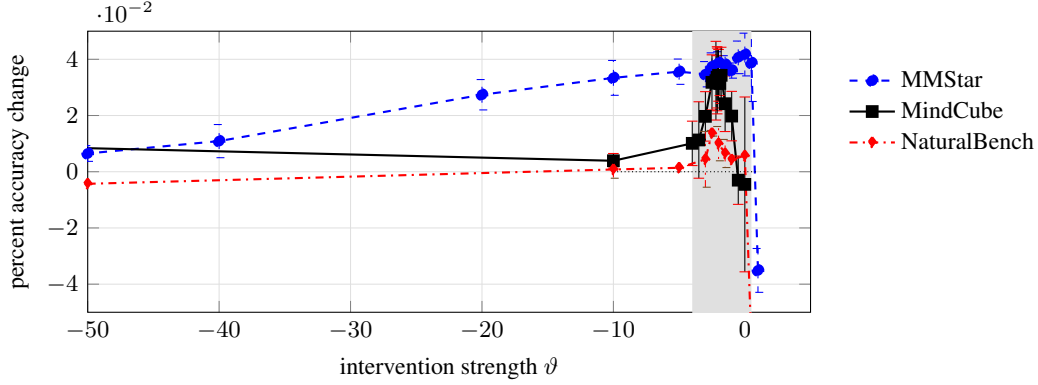

\paragraph{Benchmarks and setup.}
We evaluate performance under the same intervention sweep across MMStar,
MindCube, and NaturalBench. Each benchmark emphasizes different aspects of
vision--language performance, including reasoning, compositionality, and general
visual understanding. Unlike the core intervention sweeps, which use fixed random
seeds for replicability where supported, these benchmark evaluations are executed
through the third-party LMMS-Eval stack. We therefore replicate the benchmark
runs multiple times under independent vLLM initializations of the same
Qwen3-VL-8B model and report standard deviations across those runs as the error
bars in \Cref{fig:bench-dose-response-updated}. These LMMS-Eval benchmark data were generated on an NVIDIA GB10 system with
128~GB unified memory, using NVIDIA-SMI/driver 580.142 and CUDA 13.0, with GPU
memory utilization set to 0.3; the supplemental run manifest reports the
corresponding wall-clock/GPU-hour requirements for the MMStar, MindCube, and
NaturalBench sweeps.

\paragraph{Observed regime patterns.}
\Cref{fig:bench-dose-response-updated} shows that all three benchmarks exhibit
structured response curves under increasing intervention strength $\vartheta$.

MMStar remains elevated across a relatively wide range of $\vartheta$, suggesting
a broad plateau rather than a sharply localized interior optimum.
MindCube exhibits the clearest three-phase structure, with an identifiable
interior regime separating negligible and collapse regions.
NaturalBench shows a peak in the interior regime followed by degradation under
stronger suppression.

While the precise shape and width of the regimes vary across benchmarks, all
three display departures from monotonic degradation, consistent with the
existence of structured response regimes.

\paragraph{Interpretation.}
These results provide supporting evidence that the emergence of regime structure
is not confined to the detection-recall metric used in the main analysis.
However, because these benchmarks do not isolate the detection-derived mediator,
they do not establish causal compatibility in the sense of
Section~\ref{sec:cca}.

Accordingly, we interpret these findings as \emph{regime-consistency evidence}
rather than primary evidence of grounding-proxy alignment. Because the error bars reflect run-to-run variability in the LMMS-Eval serving
stack rather than uncertainty in the core fixed-seed intervention pipeline, they
should be read as corroborative robustness information rather than as the primary
inferential basis of the paper.

\paragraph{Relation to redistribution diagnostics.}
The regime patterns observed here are consistent with redistribution diagnostics
reported in Appendix~\ref{app:rmr_dispersion}, where intermediate intervention
strengths correspond to structured changes in attention-derived measures.
These diagnostics are presented as complementary geometric indicators rather than
causal explanations; Appendix~\ref{app:rmr_interpretation} shows that, for the 8B
hard case, $\mathrm{RMR}_{+}$ tracks the interior regime while
$\mathrm{RMR}_{-}$ and $\mathrm{RMR}_{\mathrm{obj}}$ characterize the transition to
over-suppression.

\section{Negative-Control Task Design}
\label{app:negative_control}

The primary negative control is the singleton-pair left/right spatial task in
Appendix~\ref{app:leftright_control}. Unlike the indoor/outdoor response-switch
diagnostic, this task is scored by accuracy and therefore tests whether the same
intervention sweep produces a genuine performance gain on a distinct
vision-language control task. The intended contrast is not that every control
task must degrade monotonically, but that it should not reproduce the coherent
positive interior regime observed for the detection/counting pathway.

We retain the indoor/outdoor experiment only as an auxiliary instability
diagnostic: its response-switch rate measures whether answers change relative to
baseline, not whether performance improves.

\subsection{Singleton-Pair Left/Right Spatial Control}
\label{app:leftright_control}

The singleton-pair left/right task is used as a broader spatial negative control.
Its purpose is not to instantiate the main detection-derived mediator, but to test
whether FLIP produces a reproducible bounded interior optimum on a distinct
vision-language task that depends on coarse relational judgment rather than full
localization/counting structure.

\paragraph{Dataset construction.}
The dataset is built from MS COCO val2017 in clustered-by-image form so that
downstream analysis can still use image-level cluster-robust inference. For a
candidate image to qualify, it must satisfy the same image-centric filtering logic
used in the main clustered construction, including minimum category-richness
requirements. Candidate left/right pairs are then formed only from
\emph{singleton categories}: the image must contain exactly one instance of
object$_1$ and exactly one instance of object$_2$. Additional filters require
sufficient horizontal separation between the two objects and may also impose
minimum area and maximum-overlap constraints. By default, at most one pair is
sampled per image to avoid overweighting visually busy scenes.

\paragraph{Prompt and label construction.}
Each retained pair yields a yes/no spatial-relation question of the form
\texttt{\small Is the <object1> to the left of the <object2>? Answer only yes or no.}
To maintain near-balanced labels, prompts are generated by alternating the true
ordering and the swapped ordering across qualifying pairs, producing an overall
dataset that is approximately $50/50$ yes/no.

\paragraph{Querying.}
The left/right control is queried directly from the stored text field rather than
being rebuilt from extracted object names at inference time. The querying
infrastructure otherwise matches the local chat-completions pipeline used for the
other intervention experiments, including image-path resolution, payload
construction, response extraction, and threshold bookkeeping.

\paragraph{Interpretive role.}
Because this control uses unambiguous singleton pairs, errors are less likely to
come from multiplicity or ambiguous reference than in general localization
prompts. At the same time, it does not instantiate the detection-to-counting
mediator used in the main analysis. We therefore use $\dAcc$ on this task as the
performance-based negative control: it tests whether the intervention produces a
comparable positive interior regime on a distinct spatial judgment, rather than
merely causing answer switching.

\section{Baseline Centering, Paired Designs, and Heterogeneity}
\label{app:centering}

This subsection clarifies the role of baseline-centered recall in the total-effect
model and its relationship to an equivalent paired-difference formulation under
image-clustered inference.

\paragraph{Point-estimate equivalence.}
Let $Y_i(\vartheta)$ denote the detection-side mediator for unit $i$ under
intervention strength $\vartheta$, with baseline $Y_i(-\infty)$ corresponding to the
no-intervention condition. In the present study, each image--query pair is evaluated
under both conditions, so the same units appear in both arms.

In a binary-treatment regression with an intercept, the coefficient on the treatment
indicator equals the difference in sample means:
\begin{equation}
\widehat{\alpha}_1
=
\overline{Y}(\vartheta)-\overline{Y}(-\infty).
\end{equation}
If instead we define per-sample paired differences
\begin{equation}
\delta_i = Y_i(\vartheta)-Y_i(-\infty),
\end{equation}
then the sample mean satisfies
\begin{equation}
\overline{\delta}
=
\frac{1}{N}\sum_{i=1}^N \delta_i
=
\overline{Y}(\vartheta)-\overline{Y}(-\infty).
\end{equation}
At the population level, the same equality follows from linearity of expectation:
\begin{equation}
\mathbb{E}[Y_i(\vartheta)-Y_i(-\infty)]
=
\mathbb{E}[Y_i(\vartheta)]-\mathbb{E}[Y_i(-\infty)].
\end{equation}
Thus, when both conditions are evaluated on the same set of units, global centering
and paired subtraction target the same estimand and yield the same point estimate.

\paragraph{Clustered inference for the paired design.}
The equivalence above concerns the treatment contrast, not a claim that baseline
centering is more conservative than paired subtraction. Because the pooled
Gaussian GLM uses image-level cluster-robust standard errors, baseline and
intervention observations from the same image contribute jointly to the cluster
score. The resulting variance estimate therefore accounts for the within-cluster
covariance induced by evaluating matched image--query units under both conditions.
Under the balanced paired design used here, a pooled treatment-indicator
regression with image-clustered standard errors and an equivalent
paired-difference regression with image-clustered standard errors target the same
large-sample uncertainty for the treatment contrast, up to finite-sample and
implementation differences. We use the baseline-centered pooled specification
because it keeps the reported $\Delta R_{50}$ scale aligned with the dose-response
figures, mediation summaries, and stratified GLM table, not because it is
inherently more conservative.

\paragraph{What this does and does not resolve about heterogeneity.}
The equivalence above establishes identical point estimates for the pooled treatment
contrast, but it does not by itself rule out heterogeneous effects across subgroups.
We therefore examine one interpretable source of heterogeneity directly: instance
count. The updated Qwen3-VL-4B stratification shows that the small-count stratum
closely mirrors the pooled pattern across the interior regime. The large-count
stratum also remains positive and significant over the main interior window, but
its gains are smaller and attenuate earlier, becoming non-significant at
$\vartheta=0.5$. The over-suppression reversal at $\vartheta=1.0$ is significant
in both strata, indicating that the regime structure---interior improvement
followed by collapse under strong intervention---is preserved across this
subdivision.

These findings argue against the concern that the main qualitative conclusions are an
artifact of aggregation along this dimension. If global centering were masking strong
opposing subgroup effects, one would expect sign instability across strata; instead,
both strata preserve the positive interior direction and both exhibit collapse under
over-suppression. The heterogeneity is primarily in effect size and attenuation:
few-object scenes show stronger and more persistent gains, while many-object scenes
weaken earlier. At the same time, this analysis addresses heterogeneity only along
one interpretable axis. It does not exclude the possibility that other sources of
heterogeneity---such as object category, occlusion, or image difficulty---could
modulate effect size. A broader distributional heterogeneity analysis remains an
important direction for future work.

\section{Regression Specifications and Robust Inference}
\label{app:glm}

We estimate:
\begin{align}
\dRfifty &= \alpha_0 + \alpha_1 \Iflip + \varepsilon,\\
\log \mathbb{E}[\ecount \mid \Iflip,\dRfifty]
&= \gamma_0 + \gamma_1 \Iflip + \gamma_2 \dRfifty.
\end{align}
Standard errors are clustered at the image level to account for dependence among
object queries from the same image and for the covariance between matched baseline
and intervention observations within each image cluster. The first equation is fit
as a Gaussian GLM with image-clustered SEs and supplies the $a$-term; the second
is fit as a Poisson GLM on strict counting responses and supplies the $b$-term.
The auxiliary ZeroInflatedPoisson model used in the evaluation code belongs only
to the two-part format-failure/tolerant-error decomposition for
$E_{\mathrm{combined}}$; it is not used for the reported product summary.

\paragraph{Cluster-robust stratified GLM.}
Table~\ref{tab:cluster_robust_glm} reports cluster-robust Gaussian GLM estimates
stratified by instance count ($\le 3$ vs.\ $>3$ objects) for the Qwen3-VL-4B
anchor model. The pooled estimate is practically small at $\vartheta=-50$, then
rises across the interior regime, remaining positive and significant through
$\vartheta=0.2$ before attenuating at $\vartheta=0.5$ and reversing sharply at
$\vartheta=1.0$ ($p<0.001$). The small-count stratum closely tracks the pooled
pattern. The large-count stratum is also positive and significant over the main
interior window, but attenuates earlier and is not significant at
$\vartheta=0.5$, consistent with stronger effects on cleaner few-object scenes
and weaker gains as visual multiplicity increases.

\begin{table*}[t]
\centering
\small
\caption{\textbf{Cluster-robust Gaussian GLM estimates of the intervention effect on
$\dRfifty$, stratified by instance count for Qwen3-VL-4B.}
Effects are reported pooled and by object-count bin; the coefficient of interest is
the slope on $\Iflip$.}
\label{tab:cluster_robust_glm}
\begin{tabular}{lcccc}
\toprule
$\vartheta$ & Setting & $\hat{\alpha}_{1}$ & SE & $p$-value \\
\midrule
$-50.0$ & Pooled & 0.0021 & 0.0010 & $<0.001$ \\
        & Small ($\leq 3$) & 0.0023 & 0.0010 & $<0.001$ \\
        & Large ($> 3$) & 0.0010 & 0.0010 & 0.075 \\
\midrule
$-5.0$ & Pooled & 0.0268 & 0.0020 & $<0.001$ \\
        & Small ($\leq 3$) & 0.0283 & 0.0020 & $<0.001$ \\
        & Large ($> 3$) & 0.0185 & 0.0030 & $<0.001$ \\
\midrule
$-2.5$ & Pooled & 0.0350 & 0.0020 & $<0.001$ \\
        & Small ($\leq 3$) & 0.0366 & 0.0030 & $<0.001$ \\
        & Large ($> 3$) & 0.0256 & 0.0040 & $<0.001$ \\
\midrule
$-2.0$ & Pooled & 0.0388 & 0.0020 & $<0.001$ \\
        & Small ($\leq 3$) & 0.0412 & 0.0030 & $<0.001$ \\
        & Large ($> 3$) & 0.0254 & 0.0040 & $<0.001$ \\
\midrule
$-1.5$  & Pooled & 0.0422 & 0.0030 & $<0.001$ \\
        & Small ($\leq 3$) & 0.0444 & 0.0030 & $<0.001$ \\
        & Large ($> 3$) & 0.0297 & 0.0040 & $<0.001$ \\
\midrule
$-1.0$  & Pooled & 0.0483 & 0.0030 & $<0.001$ \\
        & Small ($\leq 3$) & 0.0508 & 0.0030 & $<0.001$ \\
        & Large ($> 3$) & 0.0347 & 0.0050 & $<0.001$ \\
\midrule
$-0.5$  & Pooled & 0.0528 & 0.0030 & $<0.001$ \\
        & Small ($\leq 3$) & 0.0539 & 0.0030 & $<0.001$ \\
        & Large ($> 3$) & 0.0472 & 0.0050 & $<0.001$ \\
\midrule
$0.0$  & Pooled & 0.0590 & 0.0030 & $<0.001$ \\
        & Small ($\leq 3$) & 0.0606 & 0.0040 & $<0.001$ \\
        & Large ($> 3$) & 0.0500 & 0.0050 & $<0.001$ \\
\midrule
$0.1$  & Pooled & 0.0572 & 0.0030 & $<0.001$ \\
        & Small ($\leq 3$) & 0.0594 & 0.0040 & $<0.001$ \\
        & Large ($> 3$) & 0.0454 & 0.0050 & $<0.001$ \\
\midrule
$0.2$  & Pooled & 0.0576 & 0.0030 & $<0.001$ \\
        & Small ($\leq 3$) & 0.0604 & 0.0040 & $<0.001$ \\
        & Large ($> 3$) & 0.0419 & 0.0050 & $<0.001$ \\
\midrule
$0.5$  & Pooled & 0.0264 & 0.0040 & $<0.001$ \\
        & Small ($\leq 3$) & 0.0324 & 0.0040 & $<0.001$ \\
        & Large ($> 3$) & $-0.0068$ & 0.0060 & 0.247 \\
\midrule
$1.0$   & Pooled & $-0.5315$ & 0.0060 & $<0.001$ \\
        & Small ($\leq 3$) & $-0.5736$ & 0.0060 & $<0.001$ \\
        & Large ($> 3$) & $-0.2992$ & 0.0090 & $<0.001$ \\

\bottomrule
\end{tabular}
\end{table*}

\section{Operator-Comparison Baselines and Feature-Coherence Control}
\label{app:generic_controls}

This appendix separates two distinct comparisons that serve different roles in the
paper. The first is a full four-criterion replication across same-site
magnitude-bounding operators; the second is the graduated permutation control for
the flooring instantiation itself. Together they test whether the protocol
generalizes across operator choice while still isolating feature-coherence
dependence.

\subsection{Full Four-Criterion Replication Across Same-Site Operators}
\label{app:operator_baselines}

All operator-comparison baselines intervene at the same site as FLIP: the
post-normalization Final-site tensor passed to the output head.

\paragraph{Symmetric clipping.}
The symmetric-clipping baseline is
\begin{equation}
\widetilde{\mathbf{z}}^{\mathrm{clip}}
=\mathrm{clip}\!\left(\mathbf{z},\,-|\vartheta|,\,+|\vartheta|\right).
\end{equation}

\paragraph{Additive shift.}
The additive-shift baseline is a matched-mean translation:
\begin{equation}
\Psi_{\vartheta}(\mathbf z)=\mathbf z+\mu_{\vartheta}(\mathbf z)\mathbf 1,
\qquad
\mu_{\vartheta}(\mathbf z)
=
\frac{1}{nd}\sum_{i,k}(\vartheta-z_{ik})_+.
\end{equation}
That is, the hidden state is shifted uniformly by the mean flooring effect that the
corresponding flooring operator would induce at the same $\vartheta$.

\paragraph{Interpretation of operator replication.}
\Cref{fig:dose_response_operator_comparison} provides the compact operator-baseline
comparison at the level of primary-task dose--response alone, while
\Cref{fig:operator_comparison} extends the comparison to the full protocol for
same-site magnitude-bounding interventions. Across flooring, symmetric clipping,
and additive shift, the primary task exhibits a bounded positive regime under low
or no permutation. By contrast, the singleton-pair left/right control, scored by
$\dAcc$ rather than response switching, remains near baseline or drifts downward
over the non-collapse region and does not reproduce a comparable positive
interior regime. The grounding-proxy pathway is likewise altered once feature coherence
is disrupted. The operator-side result is therefore not that all operators behave
identically, but that the protocol continues to distinguish structured,
mediator-aligned regimes from generic perturbation effects across same-site
bounded interventions.

\noindent\Cref{fig:operator_comparison} shows the complete same-site
operator-side replication of the four-criterion protocol.
\input{figure/four-criterion-replication-operators}

\subsection{Graduated Permutation Control}
\label{app:graduated_permutation}

The graduated permutation control is not an alternative operator baseline. It keeps
the flooring instantiation fixed and tests whether its positive interior regime
depends on feature coherence.

Let $\mathbf z\in\mathbb R^{n\times d}$ denote the post-normalization Final-site
tensor, with permutation acting along the feature dimension. For mixing parameter
$\lambda\in[0,1]$, define a \emph{partial random permutation operator}
$S_\lambda\in\mathbb R^{d\times d}$ by
\begin{equation}
    S_\lambda = \Pi_\lambda \oplus I_{1-\lambda},
\end{equation}
where:
(i) $\Pi_\lambda$ is a random permutation acting on a randomly selected subset of
$\lceil \lambda d\rceil$ feature dimensions,
(ii) $I_{1-\lambda}$ is the identity on the remaining feature dimensions, and
(iii) $\oplus$ denotes the direct sum over the partition of
$\{1,\dots,d\}$ into permuted and unpermuted coordinates.

The graduated permutation control is then defined by
\begin{equation}
    \widetilde{\mathbf z}^{(\lambda)}
    =
    \max(\mathbf z S_\lambda,\vartheta),
\end{equation}
where the maximum is taken element-wise and no inverse permutation is applied.
This omission is deliberate: because elementwise flooring is permutation
equivariant, an inverse-restored construction would satisfy
\[
\max(\mathbf z S_\lambda,\vartheta)S_\lambda^{-1}
=
\max(\mathbf z,\vartheta),
\]
and would therefore collapse back to standard flooring rather than stress
feature--output-head alignment. Equivalently, a fraction $\lambda$ of feature
dimensions is permuted \emph{without inverse restoration} before flooring, while
the remaining $(1-\lambda)$ fraction is left unchanged.

This construction has clean boundary behavior:
\begin{align}
    \widetilde{\mathbf z}^{(0)}
    &= \max(\mathbf z I,\vartheta)
     = \max(\mathbf z,\vartheta),
    \\
    \widetilde{\mathbf z}^{(1)}
    &= \max(\mathbf z P,\vartheta),
\end{align}
where $P$ is a full random permutation on all $d$ feature dimensions. Thus
$\lambda=0$ recovers standard flooring, while larger $\lambda$ values progressively
destroy feature alignment by increasing the fraction of permuted coordinates.

For all reported permutation-control curves, baseline centering uses the common unpermuted no-intervention run. Let $\Rfifty^{(\lambda)}(\vartheta)$ denote recall after applying $\widetilde{\mathbf z}^{(\lambda)} = \max(\mathbf zS_\lambda, \vartheta)$. We plot
\[
\Delta R^{(\lambda)}_{50}(\vartheta)
=
R^{(\lambda)}_{50}(\vartheta)-R^{(0)}_{50}(-\infty),
\]
not $R^{(\lambda)}_{50}(\vartheta)-R^{(\lambda)}_{50}(-\infty)$. The $a$-term in the corresponding $\mathrm{IRR}_{\mathrm{indirect}}$ panels uses the same common-baseline $\Delta R^{(\lambda)}_{50}(\vartheta)$. Thus any degradation caused by permutation at $\vartheta=-\infty$ remains visible in the plotted level rather than being removed by $\lambda$-specific re-centering.

\Cref{fig:graduated_permutation_control} therefore tests whether the unpermuted
interior window survives feature-coordinate misalignment. Relative to
$\lambda=0$, all nonzero $\lambda$ settings lose the positive regime earlier and
enter collapse sooner, although the ordering among nonzero $\lambda$ values is not
strictly monotonic. The key contrast is between $\lambda=0$ and $\lambda>0$:
disrupting feature coherence destabilizes the regime-relevant window even when
the intervention class is otherwise unchanged.

\begin{figure*}[t]
\centering
\begin{tikzpicture}
\begin{axis}[
    width=\textwidth, height=8cm,
    xlabel={$\vartheta$},
    ylabel={$\dRfifty$ (relative to baseline)},
    xmin=-50, xmax=4, ymin=-0.6, ymax=0.1,
    grid=both, major grid style={draw=gray!25}, minor grid style={draw=gray!15},
    legend pos=south west, legend cell align=left,
    tick label style={font=\small}, label style={font=\small},
    legend style={font=\small, fill=white, draw=none}, line width=1.3pt, clip=true
]
\addplot[forget plot, name path=flip upper, draw=none] coordinates {
(-50,0.003)(-5,0.031)(-4,0.034)
(-2.5,0.039)(-2,0.044)(-1.5,0.047)(-1,0.054)(-0.5,0.059)
(-0.3,0.064)(-0.2,0.064)(0,0.065)(0.1,0.064)
(0.2,0.064)(0.4,0.049)(0.5,0.033)(1,-0.52)
};
\addplot[forget plot, name path=flip lower, draw=none] coordinates {
(-50,0.001)(-5,0.023)(-4,0.025)
(-2.5,0.030)(-2,0.034)(-1.5,0.037)(-1,0.043)(-0.5,0.047)
(-0.3,0.052)(-0.2,0.052)(0,0.053)(0.1,0.051)
(0.2,0.051)(0.4,0.036)(0.5,0.019)(1,-0.543)
};
\addplot[forget plot, blue!30, draw=none] fill between[of=flip upper and flip lower];
\addplot[forget plot, color=blue] coordinates {
(-50,0.002079231)(-5,0.027)(-4,0.029615497)
(-2.5,0.03495375)(-2,0.039)(-1.5,0.042174846)(-1,0.048319786)(-0.5,0.052826884)
(-0.3,0.057766647)(-0.2,0.057824289)(0,0.059)(0.1,0.057212731)
(0.2,0.057591177)(0.4,0.042474436)(0.5,0.026377262)(1,-0.531480377)
};
\addplot[forget plot, name path=clip upper, draw=none] coordinates {
(-50,0.004)(-5,0.029)(-4,0.033)
(-2.5,0.029)(-2,0.021)(-1.5,0.003)(-1,-0.014)(-0.5,-0.041)
(-0.3,-0.229)(-0.2,-0.479)(0,-0.521)(0.1,-0.52)
(0.2,-0.516)(0.4,-0.064)(0.5,-0.041)(1,-0.014)
};
\addplot[forget plot, name path=clip lower, draw=none] coordinates {
(-50,0.000)(-5,0.021)(-4,0.023)
(-2.5,0.019)(-2,0.011)(-1.5,-0.008)(-1,-0.026)(-0.5,-0.055)
(-0.3,-0.249)(-0.2,-0.502)(0,-0.544)(0.1,-0.543)
(0.2,-0.539)(0.4,-0.080)(0.5,-0.055)(1,-0.026)
};
\addplot[forget plot, red!30, draw=none] fill between[of=clip upper and clip lower];
\addplot[forget plot, color=red] coordinates {
(-50,0.001701982)(-5,0.024993348)(-4,0.028092798)
(-2.5,0.024001311)(-2,0.016156107)(-1.5,-0.002908162)(-1,-0.019968474)(-0.5,-0.048140845)
(-0.3,-0.238988732)(-0.2,-0.490611456)(0,-0.532121403)(0.1,-0.531655109)
(0.2,-0.527454499)(0.4,-0.072052253)(0.5,-0.048140845)(1,-0.019968474)
};
\addplot[forget plot, name path=shift upper, draw=none] coordinates {
(-50,0.005)(-5,0.008)(-4,0.01)
(-2.5,0.017)(-2,0.020)(-1.5,0.024)(-1,0.028)(-0.5,0.034)
(-0.3,0.038)(-0.2,0.038)(0,0.039)
(0.2,0.043)(0.5,0.044)(0.8,0.045)(1,0.045)(1.5,0.038)(2,0.017)(2.5,-0.026)(3,-0.076)(4,-0.513)
};
\addplot[forget plot, name path=shift lower, draw=none] coordinates {
(-50,-0.005)(-5,-0.002)(-4,0.0)
(-2.5,0.007)(-2,0.010)(-1.5,0.014)(-1,0.017)(-0.5,0.024)
(-0.3,0.027)(-0.2,0.027)(0,0.028)
(0.2,0.032)(0.5,0.032)(0.8,0.032)(1,0.032)(1.5,0.026)(2,0.004)(2.5,-0.040)(3,-0.091)(4,-0.536)
};
\addplot[forget plot, green!35, draw=none] fill between[of=shift upper and shift lower];
\addplot[forget plot, color=green!50!black] coordinates {
(-50,-0.000171849)(-5,0.003002437)(-4,0.005353701)
(-2.5,0.012154467)(-2,0.014804917)(-1.5,0.019428117)(-1,0.022551586)(-0.5,0.029107728)
(-0.3,0.032471972)(-0.2,0.032352764)(0,0.03380042)
(0.2,0.037669084)(0.5,0.038045478)(0.8,0.038581013)(1,0.038642287)(1.5,0.032012695)(2,0.01071388)(2.5,-0.032541796)(3,-0.08305451)(4,-0.524816136)
};
\addlegendimage{color=blue, line width=1.3pt}
\addlegendentry{Elementwise flooring}
\addlegendimage{color=red, line width=1.3pt}
\addlegendentry{Symmetric clip}
\addlegendimage{color=green!50!black, line width=1.3pt}
\addlegendentry{Additive shift}
\end{axis}
\end{tikzpicture}
\caption{\textbf{Dose--response comparison across same-site operator baselines.}
$\dRfifty$ under flooring, symmetric clipping, and additive shift, with 95\%
confidence intervals. All three bounded interventions exhibit a bounded interior
regime, but they differ in strength and collapse profile: flooring reaches the
largest interior peak, symmetric clipping collapses earliest, and additive shift
remains positive over a broader range before delayed collapse. The paper's
specificity claim is therefore narrower than operator uniqueness: curve shape
alone is insufficient, and the decisive tests are mediator alignment and
feature-coherence dependence.}
\label{fig:dose_response_operator_comparison}
\end{figure*}

\subsection{Related Operator Intuition: Symmetric Clipping and Additive Shift}
\label{app:related_operator_intuition}

The main text develops the output-head alignment baseline for flooring because
FLIP is the primary intervention. Symmetric clipping and additive shift are used
as same-site operator-comparison baselines. The relevant comparison is not that
these operators are theoretically identical, but that each induces a displacement
at the same logit-facing representation.

\paragraph{Symmetric clipping.}
The symmetric-clipping baseline is
\[
\widetilde{\mathbf z}^{\mathrm{clip}}
=
\mathrm{clip}(\mathbf z,-|\vartheta|,+|\vartheta|).
\]
It induces a displacement
\[
\Delta^{\mathrm{clip}}_{\vartheta}
=
\widetilde{\mathbf z}^{\mathrm{clip}}-\mathbf z .
\]
As $|\vartheta|$ shrinks along the sweep, more coordinates are clipped and the
induced displacement becomes broader. Any useful effect must therefore arise from
alignment between $\Delta^{\mathrm{clip}}_{\vartheta}$ and task-relevant
output-head directions, not from clipping magnitude alone.

\paragraph{Additive shift.}
The additive-shift baseline applies
\[
\Psi_{\vartheta}(\mathbf z)=\mathbf z+\mu_{\vartheta}(\mathbf z)\mathbf 1,
\qquad
\mu_{\vartheta}(\mathbf z)
=
\frac{1}{nd}\sum_{i,k}(\vartheta-z_{ik})_+ .
\]
Its displacement is global rather than support-local:
\[
\Delta^{\mathrm{shift}}_{\vartheta}
=
\mu_{\vartheta}(\mathbf z)\mathbf 1 .
\]
For fixed $\mathbf z$, $\mu_{\vartheta}(\mathbf z)$ is non-decreasing in
$\vartheta$, so the shift magnitude grows along the sweep. As with flooring and
clipping, a useful effect requires the induced displacement to project favorably
onto task-relevant output-head directions.

\paragraph{Interpretive role.}
The operator comparison therefore asks whether regime structure and
feature-coherence dependence survive changes in the displacement geometry. A regime-shaped curve alone is not sufficient; the key question is whether the
effect remains mediator-aligned, coherence-dependent, localized to the
logit-facing site, and absent from negative-control tasks.

\section{Subsample Stability of the Interior Regime}
\label{app:bootstrap}

This appendix tests whether the interior regime is stable under reductions in
effective sample size and under the finite granularity of the $\vartheta$ sweep.
Because scanning many $\vartheta$ values can otherwise overstate small peaks, we
use the grid only to define a descriptive FWHM regime window and require the
resulting window to remain visible under image-cluster subsampling rather than
claiming a separate confirmatory discovery at each grid point.

The subsampling procedure adds an argument \texttt{--n-subsamples} $N$
(default $1$). Each subsample uses seed
$\texttt{base\_seed} + s_{\mathrm{idx}}$, making the full subsampling pipeline
reproducible. For each subsample, we draw an independent permutation of the
image list without replacement. Fractions within a given subsample are then
formed cumulatively from that single permuted ordering, so that
$10\% \subset 20\% \subset 40\% \subset 60\% \subset 80\% \subset 100\%$ within the same
subsample.

\Cref{fig:bootstrap} shows that the qualitative regime is preserved across
fractions, the within-fraction means track the full-sample curve closely, and
variability decreases with scale. This supports the interpretation that the
interior regime is a stable signal rather than a small-sample artifact. These
image-cluster resampling results also serve as a robustness check on the
large-sample clustered-inference interpretation used for the reported 95\%
confidence intervals.

\begin{figure*}
\centering
\begin{tikzpicture}

\begin{groupplot}[
    group style={
        group size=3 by 2,
        horizontal sep=1.2cm,
        vertical sep=2cm
    },
    width=0.32\textwidth,
    height=0.28\textwidth,
    xlabel style={font=\small},
    ylabel style={font=\small},
    tick label style={font=\small},
    title style={font=\small},
]

\nextgroupplot[
    title={(a) $\vartheta = -0.5$},
    xlabel={Fraction (\%)},
    xlabel style={yshift=6pt},
    ylabel={$\Delta R_{50}$},
    xtick={10,20,40,60,80,100},
    xticklabel style={font=\scriptsize, rotate=45},
    ytick={0.04,0.06,0.08},
    ymin=0.04, ymax=0.08,
    clip=true,
]

\addplot+[only marks, mark=*, opacity=0.4] coordinates {
    (10, 0.0703953297173638) (10, 0.0566886854641955) (10, 0.0508724826814255) (10, 0.0537164142183956) (10, 0.04880945667236)
    (20, 0.0618018242252413) (20, 0.0517683926593504) (20, 0.0517262693733282) (20, 0.0560276284492987) (20, 0.0421778650437185)
    (40, 0.0572583731999212) (40, 0.0536033406431415) (40, 0.0531968060121591) (40, 0.0509747049350197) (40, 0.0507596193134496)
    (60, 0.0533065867082259) (60, 0.0556478081416715) (60, 0.0521492623951637) (60, 0.0527629734777314) (60, 0.0505910436368779)
    (80, 0.0537980642569933) (80, 0.0538) (80, 0.0508074883833482) (80, 0.0514058794036549) (80, 0.0538)    
};

\addplot+[
    mark=*,
    thick, 
    error bars/.cd,
    y dir=both,
    y explicit,
    error bar style={line width=1.2pt}, 
    error mark=|,
    error mark options={line width=1.2pt, mark size=3pt}, 
]
coordinates {
    (10, 5.61E-02) +- (0, 8.53E-03)
    (20, 5.27E-02) +- (0, 7.18E-03)
    (40, 5.32E-02) +- (0, 0.002623542)
    (60, 5.29E-02) +- (0, 0.001845878)
    (80, 5.30E-02) +- (0, 0.001959647)
    (100, 0.0528268844010360) +- (0, 0)
};



\nextgroupplot[
    title={(b) 10\%},
    xlabel={$\vartheta$},
    xmin=-10, xmax=1,
    ymin=-0.08, ymax=0.08,
    ytick={-0.08,0,0.08},
    clip=true,
]

\addplot+[opacity=0.2]
table[
  col sep=comma,
  x index=0,
  y index=1
]{data/10_sample1.csv};
\addplot+[opacity=0.2]
table[
  col sep=comma,
  x index=0,
  y index=1
]{data/10_sample2.csv};
\addplot+[opacity=0.2]
table[
  col sep=comma,
  x index=0,
  y index=1
]{data/10_sample3.csv};
\addplot+[opacity=0.2]
table[
  col sep=comma,
  x index=0,
  y index=1
]{data/10_sample4.csv};
\addplot+[opacity=0.2]
table[
  col sep=comma,
  x index=0,
  y index=1
]{data/10_sample5.csv};

\addplot+[very thick,solid,red, no marks]
table[x index=0, y index=1, col sep=comma]
{data/10_mean.csv};
\addplot+[very thick,densely dashed,black, no marks, opacity=0.7]
table[x index=0, y index=1, col sep=comma]
{data/100_mean.csv};

\addplot[densely dotted] coordinates {(-10,0) (1,0)};
\addplot[densely dotted] coordinates {(-0.5,-0.1) (-0.5,0.1)};

\nextgroupplot[
    title={(c) 20\%},
    xlabel={$\vartheta$},
    xmin=-10, xmax=1,
    ymin=-0.08, ymax=0.08,
    ytick={-0.08,0,0.08},
    clip=true,
]

\addplot+[opacity=0.2]
table[x index=0, y index=1, col sep=comma]
{data/20_sample1.csv};
\addplot+[opacity=0.2]
table[x index=0, y index=1, col sep=comma]
{data/20_sample2.csv};
\addplot+[opacity=0.2]
table[x index=0, y index=1, col sep=comma]
{data/20_sample3.csv};
\addplot+[opacity=0.2]
table[x index=0, y index=1, col sep=comma]
{data/20_sample4.csv};
\addplot+[opacity=0.2]
table[x index=0, y index=1, col sep=comma]
{data/20_sample5.csv};

\addplot+[very thick,solid,red, no marks]
table[x index=0, y index=1, col sep=comma]
{data/20_mean.csv};
\addplot+[thick,dashed,black, no marks, opacity=0.7]
table[x index=0, y index=1, col sep=comma]
{data/100_mean.csv};

\addplot[densely dotted] coordinates {(-10,0) (1,0)};
\addplot[densely dotted] coordinates {(-0.5,-0.1) (-0.5,0.1)};

\nextgroupplot[
    title={(d) 40\%},
    xlabel={$\vartheta$},
    ylabel={$\dRfifty$},
    xmin=-10, xmax=1,
    ymin=-0.08, ymax=0.08,
    ytick={-0.08,0,0.08},
    clip=true,
]

\addplot+[opacity=0.2]
table[x index=0, y index=1, col sep=comma]
{data/40_sample1.csv};
\addplot+[opacity=0.2]
table[x index=0, y index=1, col sep=comma]
{data/40_sample2.csv};
\addplot+[opacity=0.2]
table[x index=0, y index=1, col sep=comma]
{data/40_sample3.csv};
\addplot+[opacity=0.2]
table[x index=0, y index=1, col sep=comma]
{data/40_sample4.csv};
\addplot+[opacity=0.2]
table[x index=0, y index=1, col sep=comma]
{data/40_sample5.csv};

\addplot+[very thick,solid,red, no marks]
table[x index=0, y index=1, col sep=comma]
{data/40_mean.csv};
\addplot+[thick,dashed,black, no marks, opacity=0.7]
table[x index=0, y index=1, col sep=comma]
{data/100_mean.csv};

\addplot[densely dotted] coordinates {(-10,0) (1,0)};
\addplot[densely dotted] coordinates {(-0.5,-0.1) (-0.5,0.1)};

\nextgroupplot[
    title={(e) 60\%},
    xlabel={$\vartheta$},
    xmin=-10, xmax=1,
    ymin=-0.08, ymax=0.08,
    ytick={-0.08,0,0.08},
    clip=true,
]

\addplot+[opacity=0.2]
table[x index=0, y index=1, col sep=comma]
{data/60_sample1.csv};
\addplot+[opacity=0.2]
table[x index=0, y index=1, col sep=comma]
{data/60_sample2.csv};
\addplot+[opacity=0.2]
table[x index=0, y index=1, col sep=comma]
{data/60_sample3.csv};
\addplot+[opacity=0.2]
table[x index=0, y index=1, col sep=comma]
{data/60_sample4.csv};
\addplot+[opacity=0.2]
table[x index=0, y index=1, col sep=comma]
{data/60_sample5.csv};

\addplot+[very thick,solid,red, no marks]
table[x index=0, y index=1, col sep=comma]
{data/60_mean.csv};
\addplot+[thick,dashed,black, no marks, opacity=0.7]
table[x index=0, y index=1, col sep=comma]
{data/100_mean.csv};

\addplot[densely dotted] coordinates {(-10,0) (1,0)};
\addplot[densely dotted] coordinates {(-0.5,-0.1) (-0.5,0.1)};

\nextgroupplot[
    title={(f) 80\%},
    xlabel={$\vartheta$},
    xmin=-10, xmax=1,
    ymin=-0.08, ymax=0.08,
    ytick={-0.08,0,0.08},
    clip=true,
]

\addplot+[opacity=0.2]
table[x index=0, y index=1, col sep=comma]
{data/80_sample1.csv};
\addplot+[opacity=0.2]
table[x index=0, y index=1, col sep=comma]
{data/80_sample2.csv};
\addplot+[opacity=0.2]
table[x index=0, y index=1, col sep=comma]
{data/80_sample3.csv};
\addplot+[opacity=0.2]
table[x index=0, y index=1, col sep=comma]
{data/80_sample4.csv};
\addplot+[opacity=0.2]
table[x index=0, y index=1, col sep=comma]
{data/80_sample5.csv};

\addplot+[very thick,solid,red, no marks]
table[x index=0, y index=1, col sep=comma]
{data/80_mean.csv};
\addplot+[thick,dashed,black, no marks, opacity=0.7]
table[x index=0, y index=1, col sep=comma]
{data/100_mean.csv};

\addplot[densely dotted] coordinates {(-10,0) (1,0)};
\addplot[densely dotted] coordinates {(-0.5,-0.1) (-0.5,0.1)};

\end{groupplot}

\end{tikzpicture}

\caption{\textbf{Subsample stability of the interior regime.}
\textbf{(a)} $\Delta R_{50}$ at fixed $\vartheta=-0.5$ across subsample fractions.
Blue markers show individual resamples; red markers and error bars show the
subsample mean and variability. Fractions are formed by subsampling image
clusters. The 100\% condition is the single full-dataset
estimate and therefore has no error bar. \textbf{(b--f)} Dose--response curves for
five resamples per fraction, with the red curve denoting the within-fraction mean
and the dashed black curve the full-dataset reference. The qualitative regime is
preserved across fractions, the means track the full-sample curve closely, and
variability decreases with scale, indicating that the effect is a stable signal
rather than a small-sample artifact.}
\label{fig:bootstrap}

\end{figure*}

\section{Additional Qualitative Illustration}
\label{app:qualitative_regime}

\subsection{Illustrative Regime Visualization}
\label{app:regime_visual}
\begin{figure}[H]
  \centering
  \includegraphics[
    width=\columnwidth,
    trim=5.5cm 0cm 0cm 0.5cm,
    clip
  ]{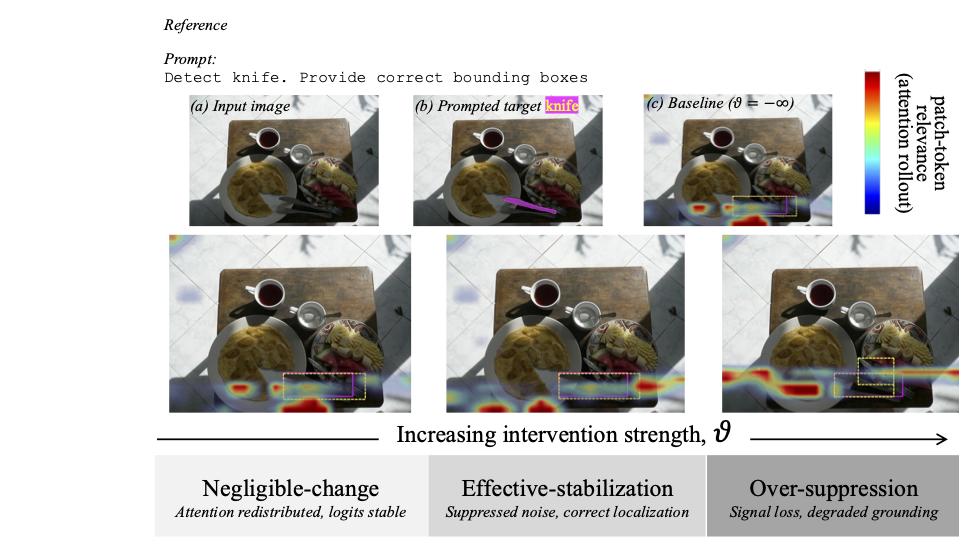}
\caption{\textbf{Illustrative regime structure under increasing intervention strength $\vartheta$.}
As $\vartheta$ increases, behavior shifts from negligible change to an interior
regime with qualitatively improved detections, and then to over-suppression.
Heatmaps show patch-token relevance from attention rollout; solid boxes denote
ground truth and dashed boxes denote model detections. Representative example only.}
  \label{fig:regime}
\end{figure}

\section{Proofs and Supporting Analysis}
\label{app:proofs}

This appendix supports the operator and output-head statements used in
\cref{sec:theory}. We state only hidden-state and pre-decoding facts; no claim is
made that decoded metrics are continuous or monotone in $\vartheta$.

\subsection{Componentwise Flooring Identity and Monotonicity}
\label{app:flooring}

For a scalar $x\in\mathbb{R}$ and threshold $\vartheta\in\mathbb{R}$, define
\[
\phi_{\vartheta}(x)=\max(x,\vartheta),
\qquad
\delta_{\vartheta}(x)=\phi_{\vartheta}(x)-x=(\vartheta-x)_+ .
\]
Applied elementwise to $\mathbf z\in\mathbb{R}^{n\times d}$,
\[
\widetilde{\mathbf z}=\phi_{\vartheta}(\mathbf z),
\qquad
\mathbf\Delta_{\vartheta}=\widetilde{\mathbf z}-\mathbf z .
\]
For each coordinate $(i,k)$,
\[
\widetilde z_{ik}=z_{ik}+(\vartheta-z_{ik})_+ .
\]

\paragraph{Proof of Proposition~\ref{prop:monotone_clamping}.}
The identity follows immediately from the definition of $\max$. If
$\vartheta_1\leq\vartheta_2$, then
\[
(\vartheta_1-z_{ik})_+ \leq (\vartheta_2-z_{ik})_+
\]
for every coordinate. Hence each entry of $\mathbf\Delta_{\vartheta}$ is
coordinatewise non-decreasing in $\vartheta$, the active support expands
monotonically, and $\|\mathbf\Delta_{\vartheta}\|_p$ is non-decreasing for any
$p\in[1,\infty]$.

\subsection{Output-Head Margin Expansion}
\label{app:output_head_margin}

For a final-token hidden state $\mathbf z_t$, logits are
\[
\ell_t = W^\top\mathbf z_t+b .
\]
Under FLIP,
\[
\widetilde{\mathbf z}_t=\mathbf z_t+\Delta_{\vartheta,t},
\qquad
\Delta_{\vartheta,t}=(\vartheta-\mathbf z_t)_+ ,
\]
so the logit change is exactly
\[
\widetilde{\ell}_t-\ell_t = W^\top\Delta_{\vartheta,t}.
\]

For token sets $Y^+$ and $Y^-$, define
\[
m(\mathbf z_t)
=
\log\sum_{y\in Y^+}\exp(w_y^\top\mathbf z_t)
-
\log\sum_{y\in Y^-}\exp(w_y^\top\mathbf z_t).
\]
The gradient of this margin is
\[
\nabla m(\mathbf z_t)
=
\bar w_{Y^+}(\mathbf z_t)-\bar w_{Y^-}(\mathbf z_t),
\]
where each $\bar w$ is the softmax-weighted average of output-head vectors within
the corresponding token set. A first-order Taylor expansion gives
\[
m(\widetilde{\mathbf z}_t)-m(\mathbf z_t)
=
\left\langle
\bar w_{Y^+}(\mathbf z_t)-\bar w_{Y^-}(\mathbf z_t),
\Delta_{\vartheta,t}
\right\rangle
+
o(\|\Delta_{\vartheta,t}\|).
\]
Thus a useful pre-decoding margin increase requires positive alignment between
the FLIP displacement and task-relevant output-head directions.

For raw decoder-layer sites, the analogous first-order effect must pass through
the remaining computation:
\[
\widetilde{\ell}-\ell \approx J_{\ell\to\mathrm{logits}}\Delta^{(\ell)}_{\vartheta}.
\]
The raw-layer analysis therefore tests whether such propagated alignment
reproduces the signature observed at the normalized logit-facing representation.

\section{FLIP versus Prompt-Based Control}
\label{app:prompt_vs_flip}

Prompt engineering operates exclusively at the level of the input distribution and
relies on the model's learned priors to steer generation. By contrast, FLIP is an inference-time intervention applied
after tokenization and cross-modal fusion, operating directly on the
post-normalization Final-site hidden state before logit computation. FLIP
does not introduce new linguistic content, cannot inject information absent from the
model's internal representations, and fails gracefully when evidence is weak or
over-suppressed---unlike prompt-based control.

\section{Model Choice, Scope, and Claims}
\label{app:scope_model}

The core detection/counting experiments are instantiated on Qwen3-VL-4B-Instruct to
enable controlled, auditable inference-time patching. Claims for the mediator-based regime analysis are therefore scoped to: (i) this
model family instance, (ii) the post-normalization Final intervention site, and
(iii) the evaluation protocol. The
contribution is the \emph{methodology}---regime-structured intervention sweeps, a
mediator-based probe, and redistribution diagnostics. Cross-architecture validation
is reported in Appendix~\ref{app:cross_architecture}.





\section{Auxiliary Instability Diagnostic: Global Scene Classification}
\label{app:sanity}

For the indoors/outdoors diagnostic, answer stability is quantified via a
\emph{baseline-referenced response-switch rate}. Let $y_i^{(-\infty)}$ and
$y_i^{(\vartheta)}$ denote the baseline and intervention outputs for sample $i$.
Outputs are canonicalized before comparison by lowercasing, extracting the first
alphabetic token, and mapping surface variants to the binary label set
\texttt{indoors}/\texttt{outdoors}
(e.g., \texttt{indoor}, \texttt{indoors}, \texttt{indo} $\mapsto$
\texttt{indoors}; \texttt{outdoor}, \texttt{outdoors}, \texttt{out} $\mapsto$
\texttt{outdoors}). Defining $c(\cdot)$ as this canonicalization map, the
response-switch rate at strength $\vartheta$ is
\begin{equation}
\mathrm{RSR}(\vartheta)
=
\frac{1}{N}\sum_{i=1}^{N}
\mathbbm{1}\!\left\{
c\!\left(y_i^{(\vartheta)}\right)\neq c\!\left(y_i^{(-\infty)}\right)
\right\}.
\end{equation}
This metric measures response instability relative to baseline rather than task
accuracy.


\end{document}